\documentclass[a4paper,fleqn]{cas-dc}

\usepackage[authoryear,longnamesfirst]{natbib}

\usepackage{amsmath,amssymb,amsfonts}
\usepackage{algorithmic}
\usepackage{graphicx}
\usepackage{algorithm,algorithmic}
\usepackage{textcomp}

\usepackage{amsmath,amsfonts}
\usepackage{algorithmic}
\usepackage{algorithm}
\usepackage{array}
\usepackage[caption=false,font=normalsize,labelfont=sf,textfont=sf]{subfig}
\usepackage{textcomp}
\usepackage{stfloats}
\usepackage{url}
\usepackage{verbatim}
\usepackage{graphicx}
\usepackage{amsmath}
\usepackage{booktabs}
\usepackage{amssymb}
\usepackage{multirow}
\usepackage{makecell}
\usepackage{booktabs}
\usepackage{multirow}
\usepackage{colortbl}
\usepackage[table]{xcolor}
\usepackage{makecell}  
\usepackage{stfloats}
\newcommand{\std}[1]{\(\scriptstyle \pm #1\)}

\def\tsc#1{\csdef{#1}{\textsc{\lowercase{#1}}\xspace}}
\tsc{WGM}
\tsc{QE}

\begin{document}
\let\WriteBookmarks\relax
\def\floatpagepagefraction{1}
\def\textpagefraction{.001}

\shorttitle{HLBG}    

\shortauthors{Li et al.}  

\title [mode = title]{Navigating Hierarchy: Hyperbolic Learning on Brain Graphs for Disorder Diagnosis}  

\tnotemark[1] 

\tnotetext[1]{} 

%

\author[1]{Yapeng Li}



\ead{liyapeng_w@163.com}



\affiliation[1]{organization={School of Computer Science and Technology, Anhui University},
            addressline={111 Jiulong Road}, 
            city={Hefei},
            postcode={230601}, 
            country={China}}

\author[1]{Bo Jiang}

\cormark[1]

\ead{jiangbo@ahu.edu.cn}



\author[1]{Ziyan Zhang}


\ead{zhangziyanahu@163.com}



\author[1]{Dongdong Chen}


\ead{chendongdong@ahu.edu.cn}



\author[1]{Zhengzheng Tu}


\ead{zhengzhengahu@163.com}
\cormark[1]


\cortext[1]{Corresponding author}



\begin{abstract}
Functional brain networks exhibit a hierarchical organization across ROI, community, and whole-brain levels, supporting local processing, inter-community coordination, and global  integration. 
Recent studies have demonstrated that brain
community-aware modeling is beneficial for both diagnosis
and biomarker identification of brain networks. 
However, existing brain graph modeling methods often struggle to model ROI–community interactions, thereby failing to fully exploit the hierarchy across ROI, community, and whole-brain network levels.  
To address this issue, 
inspired by deep hyperbolic
learning in modeling hierarchical structures, we
propose a novel framework, termed Hyperbolic Learning on
Brain Graphs (HLBG) for brain network analysis. 
The
core idea of HLBG is to exploit the inherent hierarchical geometry of hyperbolic space to model the hierarchical
relationships among ROIs, functional communities, and the
whole-brain network, thereby learning hierarchy-aware and highly
discriminative representations for brain network data. 
Specifically, HLBG first projects representations from ROIs, communities, and whole-brain network  into the Lorentzian hyperbolic space. 
Then, the multi-level hierarchy is imposed via two geometric entailment constraints.  
In addition, we further introduce a new Graph-aware Mamba (GaMamba) model, which incorporates
topology-derived structural prompts into Mamba to capture long-range dependencies while preserving graph topological
information. 
Experiments on ABIDE-I and REST-MDD demonstrate that HLBG outperforms state-of-the-art methods and identifies disorder-relevant functional biomarkers. 
\end{abstract}










\maketitle

\section{Introduction}

Brain disorders such as autism spectrum disorder (ASD) and major depressive disorder (MDD) severely impair patients' cognitive, emotional, motor, and social functions, imposing a substantial burden on families and society~~\citep{background0,background1,QIU2025103664}. Accurate and timely diagnosis is crucial for effective treatment and improved long-term outcomes. However, disease-related alterations in the brain's intrinsic functional architecture are often subtle, posing significant challenges for precise diagnosis~~\citep{background2,ma2023multi}. Functional magnetic resonance imaging (fMRI) constitutes an effective framework for investigating such abnormalities, tracking neural activity patterns through blood oxygen level-dependent (BOLD) signals~\citep{BrainGNN}. In particular, functional connectivity (FC), computed as the correlation between BOLD signal series extracted from different regions of interest (ROIs), has been widely used to model aberrant brain connectivity patterns for disorder diagnosis~\citep{survey}. Accordingly, the brain can be represented as a graph constructed from the FC matrix, where ROIs are modeled as nodes and their functional associations as edges~\citep{BioBGT}.

\begin{figure}
\centering
\includegraphics[scale=0.40]{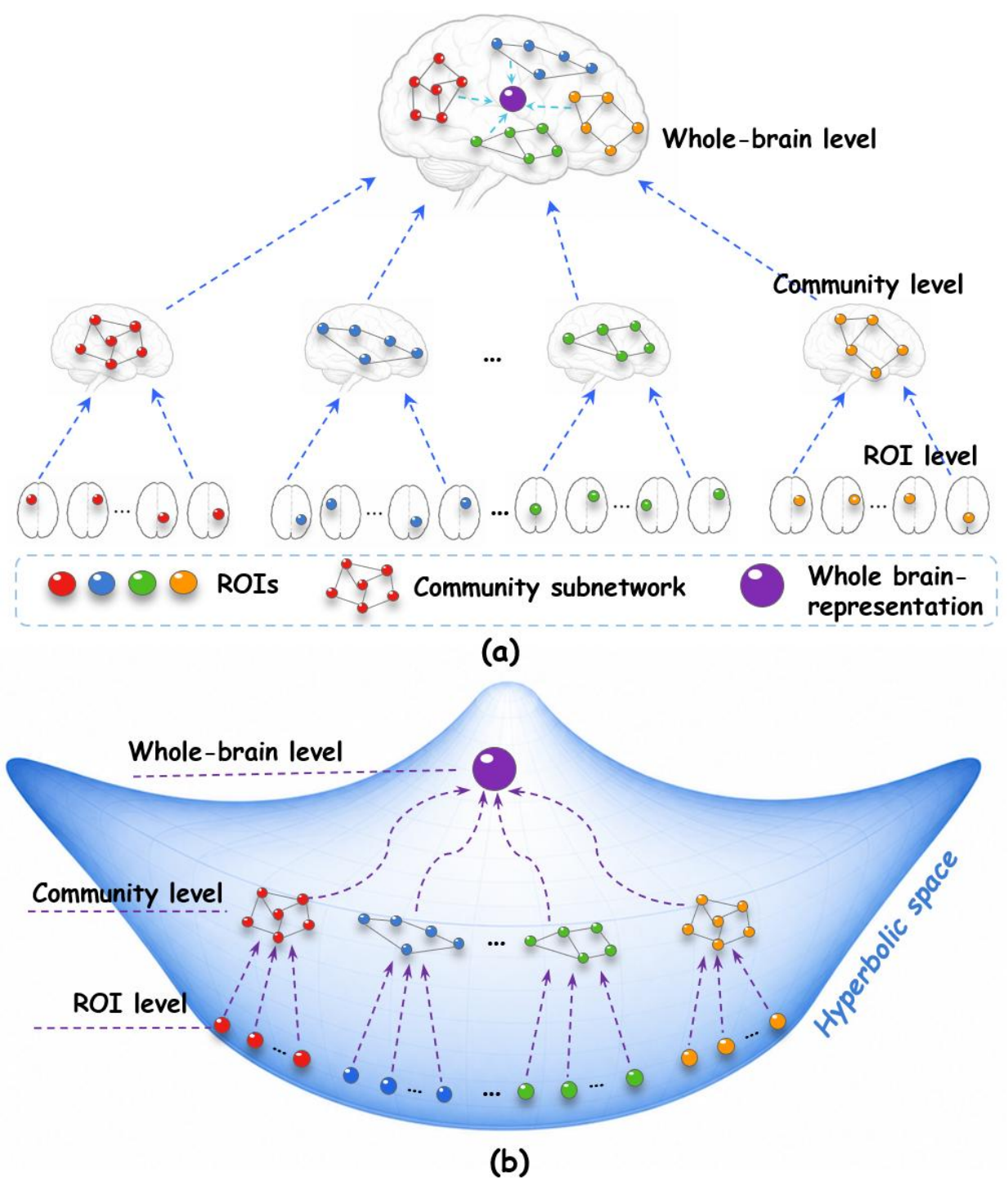}
\caption{(a) Illustration of the hierarchical organization of the brain graph across the ROI, community, and whole-brain levels. (b) Representation of the hierarchical brain network structure in hyperbolic space.} 
\label{multi_level_introduction}
\end{figure}

Graph neural networks (GNNs) have been widely used for brain network analysis tasks, supporting disease prediction and biomarker identification through graph-structured modeling~\citep{BrainGNN}.
For example, BrainGB~\citep{braingb} is a modular benchmark integrating graph construction, message passing, attention and pooling. IBGNN~\citep{IBGNN} employs edge-aware prediction and interpretable connectivity analysis. Despite these advances, conventional GNN-based methods are usually limited in modeling long-range dependencies between complex functional interactions~\citep{Trifocal_Transformer}. Graph Transformer-based methods use the global attention mechanism to learn relationships among all nodes and have shown great potential to capture global dependencies in brain network analysis. 
For instance, BrainNetTF~\citep{BrainNetTF} employs multi-head self-attention to model global interactions among brain regions. ALTER~\citep{ALTER} combines biased random walks with a Transformer framework to adaptively fuse short- and long-range information. Nevertheless, attention mechanisms may suffer from dispersed global attention and insufficient focus on local disease-relevant structures~\citep{Trifocal_Transformer,CAGT}. 
Mamba~\citep{mamba} offers an efficient way to capture long-range dependencies with linear complexity~\citep{mamba2}. Recent studies have extended it to graph data through node ordering and structure-aware fusion~\citep{GraphMamba,graphmambakdd}. However, existing Graph-Mamba models struggle to incorporate graph topology while retaining Mamba's inherent long-range modeling capability. 

Recently, some studies have demonstrated that brain community-aware modeling is beneficial for both diagnosis and interpretation of brain networks~\citep{CAGT,Com-BrainTF,ALTER}. 
By decomposing the whole-brain functional network into functionally coherent subnetworks/communities, these methods can capture fine-grained intra-community connectivity patterns and provide more interpretable disease-related biomarkers. 
For example, Com-BrainTF~\citep{Com-BrainTF} leverages community-aware local and global Transformers, together with personalized prompt tokens, to model intra- and inter-community brain connectivity patterns. CAGT~\citep{CAGT} further integrates community priors and topological information through dual-scale feature fusion and prior-guided self-attention, improving the identification of brain diseases and the discovery of biomarkers.

Despite the above advances, existing community/subgraph-aware methods often suffer from two main limitations: \textbf{First}, most existing approaches rely on Euclidean feature aggregation or attention-based fusion to integrate community-level representations. However, these strategies are inadequate for capturing the intrinsic hierarchical inter-dependencies across ROIs, meso-scale functional communities, and the macroscopic whole-brain network~\citep{CAGT,Com-BrainTF}. \textbf{Second}, although community decomposition facilitates localized subnetwork modeling, it inevitably introduces a trade‑off, i.e., the capacity to capture long‑range dependencies and cross‑community interactions among spatially distant ROIs is diminished, both of which are critical for characterizing distributed patterns of brain dysfunction.

To address these challenges, inspired by deep hyperbolic learning in modeling hierarchical structures~\citep{2025CLIPhyperbolic,1997hyperbolic,2023hyperbolic}, we propose a novel framework, termed Hyperbolic Learning on Brain Graphs (HLBG) 
for brain disorder analysis. 
The core motivation of HLBG is to exploit the inherent hierarchical geometry of hyperbolic space to model the hierarchical relationships among ROIs, functional communities, and the whole-brain network, thereby learning hierarchy-aware and discriminative representations for brain network data, as shown in Fig.~\ref{multi_level_introduction} (a).
To achieve this goal, we first construct a hierarchical brain network representation 
by organizing the global brain graph into community-specific subgraphs and arranging ROIs according to their functional-topological order~\citep{CAGT}. To explicitly characterize the hierarchical geometry of functional brain networks, we then introduce a Hierarchical Brain Representation Learning (HBRL) module, which projects ROI-, community-, and whole-brain-level representations into a unified Lorentzian hyperbolic space. Two entailment regularization losses are imposed to constrain the ROI$\rightarrow$community$\rightarrow$whole-brain hierarchical relationships, as illustrated in Fig.~\ref{multi_level_introduction} (b). This hierarchical modeling effectively captures the one-to-many associations between the whole-brain network and communities, as well as between each community and its constituent ROIs. 
In addition, we also introduce a Graph-aware Mamba model (GaMamba), which incorporates topology-derived structural prompts into Mamba to capture long-range dependencies while preserving graph topological information. 
GaMamba operates within a global-local architecture to learn complementary brain network representations: the global branch captures topology-aware representations, while multiple local branches extract community-specific features.
The updated local and global representations are adaptively fused via an attention model for brain network classification and biomarker identification. 

In summary, the main contributions of this paper are as follows:
\begin{itemize}      

\item We propose a novel Hyperbolic Learning on Brain Graphs (HLBG) scheme for brain disorder analysis, which explicitly models the hierarchical relationships among ROIs, communities, and the whole-brain network within an unified hyperbolic learning architecture. 
To the best of our knowledge, this is the first attempt to explicitly model ROI$\rightarrow$community$\rightarrow$whole-brain hierarchical relationships in the hyperbolic space for brain disorder analysis.

\item We introduce Graph-aware Mamba (GaMamba), a model designed to extract features from both community-level subgraphs and the whole-brain graph. By embedding topology-derived structural prompts into Mamba's input-dependent readout matrix, GaMamba effectively captures long-range dependencies while preserving graph structural information. 

\item Extensive experiments on the ABIDE-I and REST-MDD datasets demonstrate that the proposed HLBG outperforms state-of-the-art methods while revealing disorder-relevant functional biomarkers.
\end{itemize}

The rest of this paper is structured as follows. Section II reviews related work on brain network classification and hyperbolic learning. Section III presents the proposed HLBG framework, including hierarchical brain graph construction, brain graph embedding via GaMamba, hierarchical brain representation learning, and the overall learning objective. Section IV reports the datasets and preprocessing, implementation details, comparison results, ablation studies, hyperparameter analysis, and biomarker detection. Section V concludes this paper.

\section{Related work}

\subsection{Brain Network Classification}

Existing brain network analysis methods can be broadly categorized into two groups: 1) GNN-based methods and 2) Graph Transformer-based methods. GNN-based methods are more suited for modeling the topological structure of ROIs, enabling them to capture connectivity patterns through message-passing mechanisms~\citep{braingb,survey,THAPALIYA2025103433}. For example, Li et al.~\citep{BrainGNN} propose BrainGNN, which employs ROI-aware graph convolutional layers and ROI-selection pooling, regularized by group-level consistency and TopK pooling losses, for interpretable fMRI-based brain network analysis. 
Zhang et al.~\citep{9936686} propose a local-to-global GNN for rs-fMRI-based brain disorder classification, where a local ROI-GNN learns disease-related regional embeddings and a global subject-level GNN captures inter-subject relationships for diagnosis.
Zeng et al.~\citep{KMGCN} propose KMGCN, which integrates individual functional connectivity graphs and data-driven population graphs via multi-graph convolution, feature fusion, and graph pooling for disease classification and biomarker discovery.
Meanwhile, Graph Transformer-based approaches have recently emerged as a powerful alternative for capturing long-range dependencies and global interactions in brain graphs~\citep{BrainNetTF,GBT}. Notably, Peng et al.~\citep{GBT} combine low-rank Transformer attention with geometry-aware representation learning to enhance intra-class compactness and inter-class separation, yielding discriminative and robust functional connectome representations for autism diagnosis. Rampasek et al.~\citep{GraphGPS} propose GraphGPS, a hybrid Transformer that combines local message passing with global attention via modular positional and structural encodings to enhance scalability and expressiveness for graph representation learning. Peng et al.~\citep{BioBGT} propose BioBGT, which employs a network entanglement-based node importance encoding and a functional module-aware self-attention mechanism, together with community contrastive learning, to capture the functional segregation and integration characteristics of brain graphs and enable deep modeling of their small-world architecture.

\subsection{Learning in Hyperbolic Space}

Different from Euclidean space, hyperbolic space provides a non-Euclidean geometric setting characterized by constant negative curvature. This geometric property makes it well suited for encoding hierarchical relationships~\citep{2023hyperbolic,2025CLIPhyperbolic}. Accordingly, hyperbolic space offers a natural way to model the hierarchical organization of brain regions~\citep{baek2025mnm}. However, applying it to brain network modeling remains at an exploratory stage. For example, Jia et al.~\citep{Brain-HGCN} propose Brain-HGCN, which incorporates Lorentzian multi-head attention with signed aggregation to distinctly model positive and negative couplings and employs an intrinsic Fréchet-mean readout for intrinsic graph pooling and classification.
Baker et al.~\citep{hyperbolic_JBHI} project high-dimensional brain network representations into a low-dimensional hyperbolic manifold and characterize network organization using the radial coordinates of node embeddings, thereby examining alterations in brain network hierarchy in patients with subjective cognitive decline (SCD). Baek et al.~\citep{baek2025mnm} propose a hyperbolic space-based brain-text representation learning method, which incorporates angle-based contrastive learning, centroid regularization, and brain structural hierarchy guidance to more accurately capture multi-level relationships among brain regions. 
Different from these works, our HLBG \textbf{unifies} the modeling of hierarchical relationships across ROIs, communities, and the whole-brain network within a single hyperbolic learning framework.

\section{Methodology}
\begin{figure*}
\centering
\includegraphics[scale=0.60]{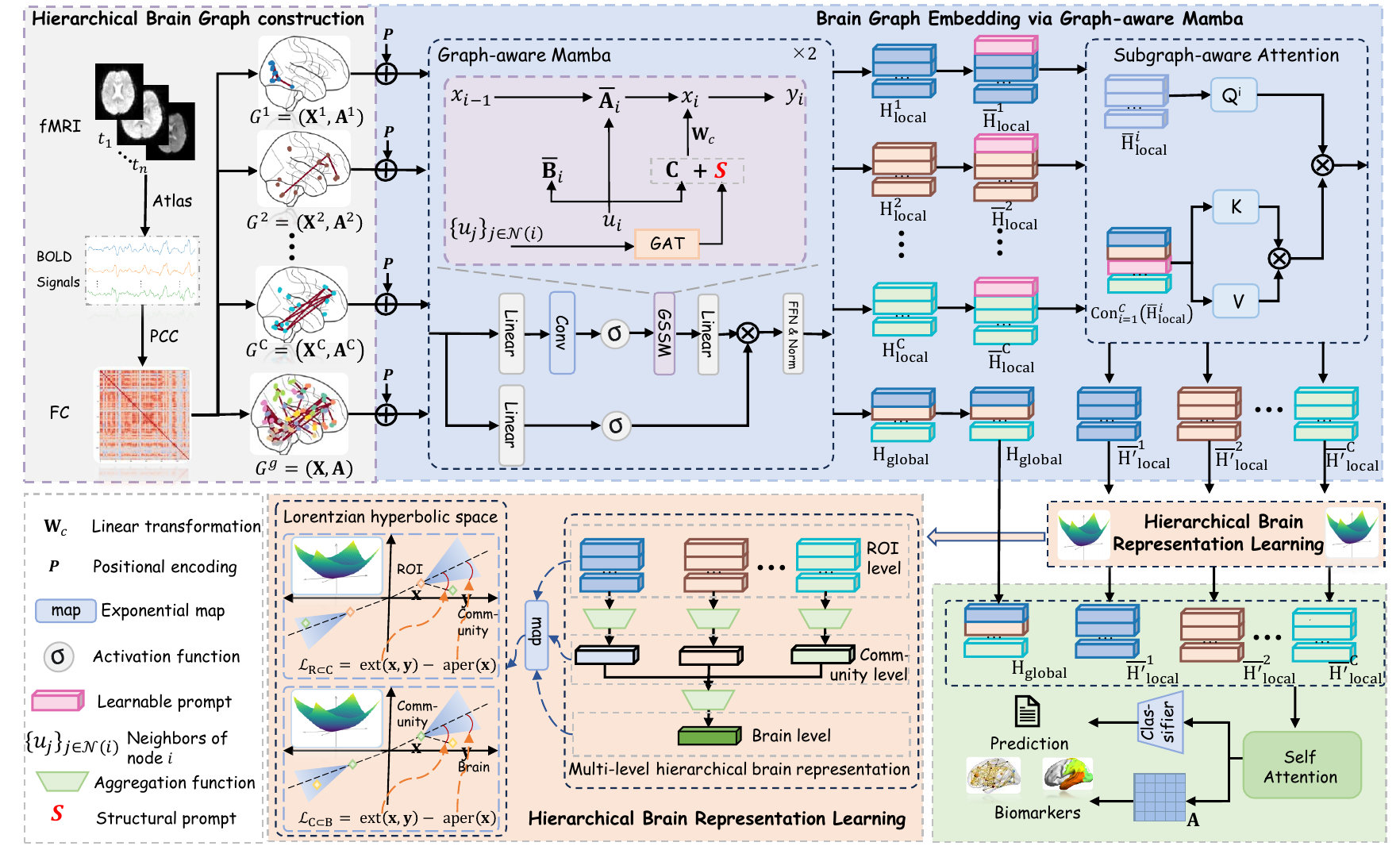}
\caption{{Overall framework of the HLBG method. It mainly includes three modules: Hierarchical Brain Graph Construction; Brain Graph Embedding  via Graph-aware Mamba; Hierarchical Brain Representation Learning.}}
\label{framework}
\end{figure*}

In this study, we propose a novel brain network analysis framework, termed \textbf{Hyperbolic Learning on Brain Graphs (HLBG)}, as illustrated in Fig.~\ref{framework}.
First, we construct a hierarchical brain graph by parcellating the global brain graph into functional community subgraphs according to the standardized functional network mapping used in~\citep{yeo7,CAGT}. 
Second, we design a multi-branch architecture leveraging brain graph feature extraction via a Graph-aware Mamba (GaMamba) module. This architecture integrates a global branch with multiple parallel local branches, which enables GaMamba to characterize topology-aware global dependencies while simultaneously encoding fine-grained intra-community features.
Third, we propose a Hierarchical Brain Representation Learning module to explicitly model the hierarchical relationships across ROI$\rightarrow$community$\rightarrow$whole-brain representations in Lorentzian hyperbolic space.
Finally, a self-attention mechanism is employed to adaptively fuse local and global representations for brain network classification and biomarker discovery.

\subsection{Hierarchical Brain Graph Construction}

To better learn hierarchical representations in the brain network, we construct a hierarchical brain graph as $\mathcal{H} = \{G^g, G^1, G^2, \dots, G^C\}$, where $G^g$ denotes the global brain graph and $G^i$ denotes the $i$-th local community-specific subgraph. 
For the global brain graph, we first partition the brain into \( N \) regions of interest (ROIs) using a standard brain atlas~\citep{yeo7,CAGT}  and compute the Pearson correlation coefficients between all pairs of ROIs to obtain a functional connectivity (FC) matrix \( \mathbf{F} \in \mathbb{R}^{N \times N} \). 
Then, we formulate the global brain graph as $G^g = (\mathbf{X},\mathbf{A})$, where the node feature matrix \( \mathbf{X}= [x_1, x_2, \dots, x_N] \) is derived directly from the FC matrix $\mathbf{F}$, with each \( x_n \in \mathbb{R}^N \) corresponding to the \( n \)-th row of \( \mathbf{X} \). The adjacency matrix \( \mathbf{A}^g \in \mathbb{R}^{N \times N}\) retains each node's  \( k \)-strongest functional connections from the FC matrix.
For each local community-specific subgraph, 
following~\citep{CAGT}, we reorganize $G^g$ into $C$ functionally homogeneous subsystems (communities). Formally, each local subgraph $G^i = (\mathbf{X}^i, \mathbf{A}^i)$ is induced from the global graph, where $\mathbf{X}^i \in \mathbb{R}^{N_i \times N}$, $\mathbf{A}^i \in \mathbb{R}^{N_i \times N_i}$ and $N_i$ denotes the number of nodes in the $i$-th community. 
Here, $\mathbf{X}^i$ and $\mathbf{A}^i$ are obtained by extracting rows and columns from $\mathbf{X}$ and $\mathbf{A}$ corresponding to the members of the $i$-th community, thereby preserving both the intrinsic community topology and the node features.

\subsection{Brain Graph Embedding via GaMamba} 

To jointly capture long-range dependencies and local topological patterns, we develop a multi-branch architecture for hierarchical brain graph embedding. Within this framework, inspired by~\citep{ICLR2025_Structure-Aware_mamba,guo2025mambairv2}, GaMamba is proposed to encode both global and community-specific representations.
The core of GaMamba lies in injecting graph structure as a structural prompt into Mamba’s input-dependent readout matrix, enabling simultaneous modeling of sequential dynamics and graph-aware representations.
  Specifically, GaMamba encodes structural priors within the state space by leveraging a Graph Attention Network (GAT)~\citep{GAT}. The GAT aggregates features across neighborhood $\mathcal{N}(i)$ to derive a normalized structural prompt, $\textbf{S}_i = \text{Norm}(\text{GAT}(\{u_j\}_{j \in \mathcal{N}(i)}))$. 
This prompt is then integrated into Mamba's output matrix \(\mathbf{C}\), an input-dependent readout parameter in the selective SSM~\citep{mamba,mamba2}, via a linear transformation with parameter \(\mathbf{W}_c\), yielding a structure-aware matrix \(\mathbf{C}^{\prime}\). \textcolor{black}{In this configuration, while hidden states continue to evolve according to discrete state-space equations, the output layer explicitly encodes the brain’s topological organization through the injected structural priors.}
Overall, the key formulation of GaMamba is as follows:
\begin{equation}
\begin{aligned}
x_{i}&=\overline{\mathbf{A}}x_{i-1}+\overline{\mathbf{B}}u_{i}, \\
\mathbf{C}^{\prime}&=\mathbf{W}_c\cdot(\mathbf{C}+\mathbf{S}),\\
y_{i}&=\mathbf{C}^{\prime}x_{i}+\mathbf{D}u_{i},
\end{aligned}
\end{equation}
where \(u_i\) denotes the input of the \(i\)-th node, 
and \(x_i\) denotes the latent state of the SSM at the \(i\)-th step. 
\(\overline{\mathbf{A}}\) is the discretized state transition matrix, 
and \(\overline{\mathbf{B}}\) is the input-to-state mapping matrix. 
\(\mathbf{D}\) is the skip connection parameter, and \(y_i\) is the final output representation at the \(i\)-th step.

Based on the proposed GaMamba module, we develop a multi-branch architecture for hierarchical brain graph embedding.
Specifically, the input brain functional graph is processed in parallel by two complementary branches to explicitly model global and local dependencies. 
\textcolor{black}{The global branch is designed to capture long-range interactions across all nodes, thereby encoding holistic topological characteristics of the brain network. 
In contrast, the local branch focuses on intra-community connectivity patterns, enabling the modeling of fine-grained regional structures.}
To enhance the representation capability of both branches, we incorporate positional encoding \(\mathbf{P}\), which is constructed by concatenating Random Walk Positional Encoding (RWPE) with normalized MNI spatial coordinates to jointly encode topological and spatial information, following~\citep{CAGT}.
\textcolor{black}{This design allows the model to preserve relative positional relationships among nodes within the graph embedding space.}
Concretely, the input node features are linearly projected and fused with the positional encoding to generate branch-specific representations as:
\begin{equation}
\begin{aligned}
{\mathbf{X}_{{local}}}&=\mathbf{W}_{local}\mathbf{X} + \mathbf{W}'_{local}\mathbf{P},\\
   {\mathbf{X}_{\text{global}}}&=\mathbf{W}_{global}\mathbf{X}  + \mathbf{W}'_{global}\mathbf{P},
\end{aligned}\end{equation}        
where $\{\mathbf{W}_{global},\mathbf{W}'_{global}, \mathbf{W}_{local}, \mathbf{W}'_{local} \}$ are four different learnable weights.
Next, we employ $C+1$ independent $l$-layer GaMamba modules to encode the global brain graph and $C$ local community-specific subgraphs as:
\begin{equation}\begin{aligned}
 \mathbf{H}_{\text{global}} &=\text{GaMamba} \left(\mathbf{{X}_{\text{global}}},\mathbf{A}\right),\\
 \mathbf{H}_{\text{local}}^{i} &=\text{GaMamba} \left(\mathbf{{X}}^{i}_{\text{local}},\mathbf{A}^i\right), \\
\end{aligned}\end{equation}
where \( \mathbf{H}_{\text{global}} \) denotes the global representation, and \( \mathbf{H}_{\text{local}}^{i} \) denotes the local representation of the \( i \)-th subgraph.

After extracting the local features of each functional subgraph, we employ a Subgraph-aware Attention (SA) module  to integrate local subgraph representations. 
Specifically, for the \(i\)-th subgraph, we concatenate its local feature 
\(\mathbf{H}_{\text{local}}^{i}\in \mathbb{R}^{N_i \times d}\) 
with a learnable initial embedding 
\(\mathbf{Z}^i\in \mathbb{R}^{N_b \times d}\), 
which serves as learnable prompt tokens to aggregate and exchange subgraph-level information, 
to obtain the prompt-augmented subgraph embedding 
\(\mathbf{\overline{H}}_{\text{local}}^{i}\in \mathbb{R}^{(N_i+N_b) \times d}\), 
i.e., 
\(\mathbf{\overline{H}}_{\text{local}}^{i}
=[\mathbf{H}_{\text{local}}^{i}\Vert\mathbf{Z}^i]\).
Subsequently, we apply cross-attention layers (CA)~\citep{subgraph_fusion} to fuse the subgraph features as: 
\begin{equation}
\begin{split}
\mathbf{Q}^i =&\mathbf{\overline{H}}_{\text{local}}^{i},
\mathbf{K} =\mathbf{V}= {\operatorname{Concat}_{i=1}^{C} \left( \mathbf{\overline{H}}_{\text{local}}^{i} \right) },\\
& \mathbf{\overline{H'}}_{\text{local}}^{i} = \text{CA}(\mathbf{Q}^i,\mathbf{K},\mathbf{V}),\\
\end{split}
\end{equation}
where \( \mathbf{\overline{H'}}_{\text{local}}^{i} \) represents the updated representations of the \(i\)-th subgraph, \(\operatorname{Concat}_{i=1}^{C}(\cdot)\) concatenates all \(C\) prompt-augmented subgraph representations along the node dimension to construct the shared key and value, and the final output \(\mathbf{H'}_{\text{local}}\) is obtained by removing the updated prompt embeddings and concatenating the ROI-level subgraph representations.

\subsection{Hierarchical Brain Representation Learning}

To enable hierarchical brain graph modeling, we propose a Hierarchical Brain Representation Learning (HBRL) module that projects multi-level brain representations into Lorentzian hyperbolic space and imposes explicit hierarchical constraints across ROIs, communities, and whole-brain levels.
Specifically, the hierarchical representation consists of ROI-level features $\mathbf{Z}^{r}\in\mathbb{R}^{N\times d}$, community-level features $\mathbf{Z}^{c}\in\mathbb{R}^{C\times d}$, and a whole-brain embedding $\mathbf{Z}^{b}$. 
The ROI-level features $\mathbf{Z}^{r}$ are initialized from $\mathbf{H'}_{\text{local}}$.
The community-level and whole-brain representations are then obtained by aggregating ROI-level features via an aggregation function $\phi(\cdot)$:
\begin{equation}
\mathbf{Z}^{c}_{i} = \phi(\{\mathbf{Z}^{r}_j|j\in\mathcal{C}_i\}), \quad
\mathbf{Z}^{b} = \phi(\mathbf{Z}^{r}),
\end{equation}
where $\mathcal{C}_i$ denotes the set of ROIs belonging to the $i$-th community. 
Taking the calculation of $\mathbf{Z}^{b}$ as an example, 
the aggregation function $\phi(\cdot)$ is implemented as an attention-based weighted summation over ROI-level features:
\begin{equation}
  \mathbf{Z}^{b}=\phi(\mathbf{Z}^{r}) = \sum_{i=1}^{N}\alpha_i\mathbf{Z}^{r}_i,\
\alpha_i=\frac{\exp\!\big(\mathbf{\omega}^\top \tanh(\mathbf{W} \mathbf{Z}^{r}_i)\big)}{\sum_{j=1}^N \exp\!\big(\mathbf{\omega}^\top \tanh(\mathbf{W} \mathbf{Z}^{r}_j)\big)} ,
\end{equation}
where \textcolor{black}{\( \mathbf{W} \in \mathbb{R}^{d_a \times d} \) and \( \mathbf{\omega} \in \mathbb{R}^{d_a} \)} are two learnable weight parameters, $d_a$ is the attention hidden dimension.
The community-level embeddings $\mathbf{Z}^{c}_{i}$ are computed in the same manner, with the aggregation restricted to nodes within each community $\mathcal{C}_i$.

Next,
we project the hierarchical representations $\mathbf{Z}^{r}$, $\mathbf{Z}^{c}$ and $\mathbf{Z}^{b}$ from Euclidean space into the Lorentzian hyperbolic space via the exponential map $\exp_0^\kappa(\cdot)$ following~\citep{2025CLIPhyperbolic}, yielding the corresponding hyperbolic embeddings 
$\tilde{\mathbf{Z}}^{r}$, $\tilde{\mathbf{Z}}^{c}$ and $\tilde{\mathbf{Z}}^{b}$. 
The exponential map is defined as:
\begin{equation}
\label{exp}
\exp_z^\kappa(v) = \cosh\!\left( \sqrt{\kappa} \|v\|_{\mathcal{L}} \right) z + \frac{\sinh\!\left( \sqrt{\kappa} \|v\|_{\mathcal{L}} \right)}{\sqrt{\kappa} \|v\|_{\mathcal{L}}} \, v,
\end{equation}
where \(\|v\|_{\mathcal L} = \sqrt{\langle v, v \rangle_{\mathcal L}}\). Let
\(\kappa > 0 \) denote the absolute value of the negative curvature; the Lorentz model has curvature \(-\kappa\), \(z\) is the base point of the exponential map.
 The bilinear form \(\langle \cdot, \cdot \rangle_{\mathcal{L}}\) denotes the Lorentzian inner product as:
\begin{equation}
\langle x, y \rangle_{\mathcal{L}} = -x_0 y_0 + \sum_{i=1}^{d} x_i y_i,
\end{equation}
where $x = (x_0, x_1, \dots, x_d)$, \(x_0\) denotes the time component and the remaining components \((x_1, \dots, x_d)\) represent the spatial components~\citep{1997hyperbolic,2023hyperbolic}. 

To explicitly model the hierarchical relationships among ROIs, communities, and the whole-brain, we define two entailment losses in Lorentzian hyperbolic space:
1) \textbf{ROI-in-Community Loss} $\mathcal{L}_{\mathrm{R\subset C}}$, which enforces that each community feature $\tilde{\mathbf{Z}}^c_i$ entails its corresponding ROI features $\{\tilde{\mathbf{Z}}^r_j|j\in\mathcal{C}_i\}$.
2) \textbf{Community-in-Brain Loss} $\mathcal{L}_{\mathrm{C\subset B}}$, which enforces that the brain feature $\tilde{\mathbf{Z}}^b$ entails all community features $\tilde{\mathbf{Z}}^c$.
Formally, the two losses can be written as:
\begin{equation}\begin{aligned}
&\mathcal{L}_{\mathrm{R\subset C}}=\frac{1}{N}\sum_{j=1}^C\sum_{i=1}^{N_j}\mathcal{L}(\tilde{c}^j,\tilde{x}_{i,j}),\\
&
\mathcal{L}_{\mathrm{C\subset B}}=\frac{1}{C}\sum_{j=1}^C\mathcal{L}(\tilde{b},\tilde{c}^{j}),
\end{aligned}\end{equation}
where \(\mathcal{L}(x, y)\) denotes the entailment loss between the parent node \(x\) and the child node \(y\).
The purpose of \(\mathcal{L}(x, y)\) is to enforce that the child node \(y\) lies within the entailment cone defined by the parent node \(x\).  
This is achieved by constraining the outer angle \(\mathrm{ext}(\mathbf{x}, \mathbf{y})\) to not exceed the half-aperture angle \(\mathrm{aper}(\mathbf{x})\):
\begin{equation}
\mathcal{L}(x, y) = \max\big(0, \, \mathrm{ext}(\mathbf{x}, \mathbf{y}) - \mathrm{aper}(\mathbf{x})\big),
\end{equation}
where the outer angle $\mathrm{ext}(\mathbf{x}, \mathbf{y})$ and the half-aperture angle $\mathrm{aper}(\mathbf{x})$ are computed as:
\begin{equation}\begin{split}
\mathrm{aper}(\mathbf{x})&=\sin^{-1}\left(\frac{2K}{\sqrt{\kappa}\left\|\mathbf{x}_{1:d}\right\|}\right),\\
\mathrm{ext}(\mathbf{x},\mathbf{y})&=\cos^{-1}\left(\frac{y_{0}+x_{0}\mathrm{~\kappa~}\langle\mathbf{x},\mathbf{y}\rangle_{\mathcal{L}}}{\|\mathbf{x}_{1:d}\|\sqrt{\left(\kappa\mathrm{~}\langle\mathbf{x},\mathbf{y}\rangle_{\mathcal{L}}\right)^2-1}}\right),\end{split}\end{equation}
where \( x_{1:d} = (x_1, \dots, x_d) \) denotes the spatial components, and \( K \) is a constant set to 0.1 here to impose a boundary condition near the origin~\citep{ganea2018hyperbolic}. 
Note that, here we explicitly construct a hierarchical structure of brain networks in Lorentzian hyperbolic space, enabling the model to learn structurally consistent, hierarchically organized, and more discriminative representations, which facilitates capturing the brain’s complex functional modularity and global topological organization.

\subsection{Fusion and Learning Loss}
We concatenate the global and local node representations to form the final node-level feature representation:
\begin{equation}
\label{12}
\begin{split}
& \mathbf{H} = \text{Concat}\left( \mathbf{\overline{H}'}_{\text{local}},\; \mathbf{{H}}_{\text{global}} \right),\\
&(\mathbf{{H}}_{\text{SA}},\textbf{A}_{att})= \text{SelfAttn}(\textbf{H}), \mathbf{{H}}_{\text{final}} =   w \cdot  \mathbf{H}_{\text{SA}} + \mathbf{H},
\end{split}\end{equation} where $\text{SelfAttn}(\cdot)$ represents the self-attention operation, $\textbf{A}_{att}$ denotes the attention weight matrix and $w$ denotes a weight coefficient.
To obtain the graph-level representation, we employ the same readout strategy as in ~\citep{BrainNetTF,CAGT}, aggregating node-level features into a graph-level embedding. Specifically, we introduce the OCRead module for structured graph-level representation learning. This module first maps node embeddings into a latent space and learns a set of orthogonal cluster centers to capture underlying functional patterns. Then, through a soft assignment mechanism, each node is assigned to these centers, and node features are weighted and aggregated according to the assignment weights to produce the final graph-level representation. Lastly, a linear layer is applied to generate the predicted output.

To simultaneously capture the hierarchical structure of brain networks and perform the downstream classification task, our model combines two types of losses: two-level entailment losses in Lorentzian hyperbolic space (local entailment and global entailment) and a supervised cross-entropy loss. The overall optimization objective is:

\begin{equation}
\label{13}
\mathcal{L} = \lambda_{\text{1}} \mathcal{L}_{\mathrm{R\subset C}} + \lambda_{\text{2}} \mathcal{L}_{\mathrm{C\subset B}} + \mathcal{L}_{\text{CE}},
\end{equation}
where \(\lambda_{\text{1}}\) and \(\lambda_{\text{2}}\) are weighting coefficients, $\mathcal{L}_{\text{CE}}$ denotes the cross-entropy loss.

\section{Experiments and Results}
\subsection{Datasets and Preprocessing}

In this study, we employ two publicly available fMRI datasets for evaluating our method: 

 \textbf{1) ABIDE-\uppercase\expandafter{\romannumeral1} dataset}  is designed to advance the neurobiological understanding of ASD by aggregating neuroimaging data from 17 international sites\footnote{\url{https://fcon\_1000.projects.nitrc.org/indi/abide/abide\_I.html}.}. We utilize the preprocessed fMRI data from the publicly released and shared ABIDE-I dataset by the Preprocessed Connectomes Project (PCP)\footnote{\url{http://preprocessed-connectomes-project.org/abide/}.}~\citep{ABIDE}, selecting fMRI from 530 normal control (NC) subjects and 505 individuals diagnosed with ASD. Detailed demographic characteristics are summarized in Table~\ref{tab:Dataset statistics}.

 \textbf{2) REST-MDD dataset} ~\citep{MDD,MDD1} from the REST-meta-MDD Project\footnote{\url{https://rfmri.org/REST-meta-MDD}.}, is a large-scale, multi-site resting-state fMRI dataset from 25 cohorts across 17 hospitals in China. To improve reproducibility and reduce methodological variability, all sites adopted a harmonized preprocessing pipeline based on DPARSF~\citep{dpabi}. We select 1,104 NC subjects and 1,276 individuals diagnosed with MDD. Detailed demographic characteristics are provided in Table~\ref{tab:Dataset statistics}.

We employ the Craddock 200 atlas~\citep{CC200} to partition the whole brain into 200 regions of interest (ROIs) across all datasets. Based on Pearson correlation coefficients (PCC)~\citep{PCC} computed among these ROIs, FC matrices were constructed. Following established practices in prior studies~\citep{yeo7,8subgraph2,CAGT}, these 200 ROIs were assigned to eight major functional sub-networks: Ventral Attention Network (VAN), Cerebellum and Subcortical Structures (CB\&SC), Visual Network (VN), Default Mode Network (DMN), Frontoparietal Network (FPN), Limbic Network (LN),  Dorsal Attention Network (DAN) and Somatomotor Network (SMN).

\subsection{Implementation Details}
All experiments are implemented in Python~3.9 with PyTorch~2.2.1 and run on an NVIDIA RTX 3090 GPU with 24~GB memory. During training, HLBG is optimized using the Adam optimizer, where the weight decay, initial learning rate, and batch size are set to $1 \times 10^{-6}$, $1 \times 10^{-4}$, and 64, respectively. The number of top-$k$ highest correlation values is set to 30, and the GaMamba module consists of two layers. The absolute curvature parameter $\kappa$ is initialized to 1.3 for ABIDE-I and 0.39 for REST-MDD, while the prompt token size $N_b$ is set to 10 and 8, respectively. For ABIDE-I, the weighting coefficients are $\lambda_1=0.4$, $\lambda_2=0.1$, and $w=0.3$; for REST-MDD, $\lambda_1=\lambda_2=0.3$ and $w=0.1$. Following~\citep{CAGT}, model performance is assessed using ten-fold cross-validation, with each metric summarized by its mean and standard deviation across the folds.

\begin{table}
\centering
\caption{Dataset statistics. (M: MALE, F: FEMALE)}
\label{tab:Dataset statistics}

\resizebox{\columnwidth}{!}{%
\begin{tabular}{c|c c|c c}
\toprule
\toprule
\textbf{Dataset}     
    & \multicolumn{2}{c|}{\textbf{ABIDE-I}} 
    & \multicolumn{2}{c}{\textbf{REST-MDD}}  \\ 
\midrule

\textbf{Class}       
    & NC & ASD 
    & NC & MDD\\ 
\midrule

\textbf{Number}      
    & 530 & 505 
    & 1104 & 1276 \\

\textbf{Gender (M/F)} 
    & 435/95 & 443/62
    & 462/642 & 463/813\\

\textbf{Age}         
    & 16.81\std{7.44} & 17.10\std{8.54}
    & 36.15\std{15.66} & 36.00\std{14.62} \\
\bottomrule

\end{tabular}%
}
\end{table}

\begin{table*}[t]
\centering
\footnotesize
\setlength{\tabcolsep}{6pt}
\renewcommand{\arraystretch}{1.25}
\caption{ Comparisons with state-of-the-art methods (\%). 
For each metric, the highest performance is indicated by a \cellcolor{cyan!10}{cyan} background, and the second-highest is \underline{underlined}.}
\label{tab:Comparison-methods}
\resizebox{0.8\textwidth}{!}{%
\begin{tabular}{cl|ccc|ccc}
\Xhline{1.2pt}
\toprule
\toprule
 & \textbf{Dataset} 
 & \multicolumn{3}{c|}{\textbf{ABIDE-I}} 
 & \multicolumn{3}{c}{\textbf{REST-MDD}}  \\
\midrule
 & \textbf{Model}
 & ACC & SEN & SPE 
 & ACC & SEN & SPE\\
\midrule

\multirow{5}{*}{\rotatebox{90}{\textbf{GNN}}}
 & GINE 
 & 63.04\std{2.90} & 65.93\std{8.41} & 64.82\std{9.81} 
 & 60.97\std{3.89} & 61.62\std{7.81} & 63.18\std{3.89}  \\

 & BrainGNN 
 & 65.12\std{3.11} & 62.87\std{13.82} & 65.07\std{14.51}
 & 65.79\std{5.01} & 64.35\std{9.81} & 34.25\std{15.21}\\

 & IBGNN
 & 66.28\std{4.49} & 65.66\std{13.63} & 66.98\std{9.40} 
 & 62.10\std{2.80} & 69.05\std{10.00} & 54.07\std{14.21}\\

 & BrainUSL
 & 70.72\std{4.09} & 71.09\std{12.98} & 68.94\std{11.53}
 & 65.00\std{1.34} & 71.49\std{8.41} & 56.59\std{10.08}\\

 & BrainIB
 & 67.34\std{2.63} & 70.94\std{8.36} & 63.57\std{7.60}
 & 62.73\std{2.42} & 68.41\std{3.71} & 56.16\std{6.35}\\
\midrule

\multirow{8}{*}{\rotatebox{90}{\textbf{GT}}}
 & vanillaTF
 & 64.20\std{3.35} & 66.40\std{6.97} & 62.47\std{10.06}
 & 61.73\std{1.25} & 64.22\std{3.17} & 52.10\std{7.35}\\

 & BrainNetTF
 & 69.22\std{3.15} & 68.69\std{5.68} & 65.13\std{4.41}
 & 65.33\std{2.18} & 67.77\std{5.89} & 62.54\std{4.55}\\

 & Com-BrainTF
 & 68.93\std{4.45} & 69.78\std{6.99} & 66.11\std{4.75}
 & 65.42\std{2.04} & 67.59\std{4.14} & 63.21\std{4.22}\\

 & ALTER
 & 71.29\std{3.76} & 72.23\std{6.21} & 66.10\std{8.09}
 & 66.93\std{2.39} & 71.70\std{4.79} & 61.32\std{3.77}\\

 & GBT
 & 70.06\std{4.96} & 73.08\std{7.73} & 66.86\std{7.73}
 & 67.04\std{2.33} & 70.25\std{6.27} & \underline{64.88\std{6.10}}\\

 & Contrasformer
 & 67.30\std{3.83} & 69.66\std{13.57} & 65.09\std{17.09}
 & 63.30\std{2.91} & 63.95\std{4.70} & 50.90\std{3.92}\\

 & GraphGPS
 & 65.30\std{2.63} & 60.61\std{6.57} & 64.12\std{7.32}
 & 61.09\std{1.32} & 65.56\std{2.08} & 53.45\std{3.62}\\

 & CAGT
 & \underline{74.01\std{3.33}} & \underline{77.83\std{8.63}} & \underline{69.60\std{9.55}}
 & \underline{68.23}\std{2.05} & 71.57\std{6.29} & 64.11\std{7.81}\\
\midrule
\multirow{3}{*}{\rotatebox{90}{\textbf{ MB}}}
 & Mamba
 & 70.04\std{4.12} & 70.43\std{10.69} & 69.03\std{4.95}
 & 67.02\std{2.28} & \underline{71.92\std{6.36}} & 61.21\std{7.26}\\

 & GraphMamba
 & 71.11\std{2.95} & 74.76\std{7.25} & 66.72\std{4.83}
 & 66.22\std{1.76} & 70.66\std{11.73} & 60.35\std{10.02}\\

\midrule

  & \cellcolor{cyan!10}Ours (HLBG)
 & \cellcolor{cyan!10}75.45\std{3.32} 
 & \cellcolor{cyan!10}79.09\std{7.59} 
 & \cellcolor{cyan!10}71.36\std{5.68}
 & \cellcolor{cyan!10}70.13\std{2.27} 
 & \cellcolor{cyan!10}73.54\std{6.69} 
 & \cellcolor{cyan!10}65.78\std{6.99}
\\
\bottomrule

\end{tabular}
}
\end{table*}

\subsection{Comparison Results}

To evaluate the effectiveness of our HLBG method, we employ three metrics: Accuracy (ACC), Specificity (SPE), and Sensitivity (SEN). We benchmark our approach against three classes of models: (1) GNN-based methods (GNN): GINE~\citep{GINE}, BrainGNN~\citep{BrainGNN},  IBGNN~\citep{IBGNN}, BrainUSL~\citep{BrianUSL}, and  BrainIB~\citep{BrainIB}; (2) Graph Transformer-based methods (GT): Transformer (vanillaTF)~\citep{transformer}, BrainNetTF~\citep{BrainNetTF}, Com-BrainTF~\citep{Com-BrainTF}, ALTER~\citep{ALTER}, GBT~\citep{GBT}, Contrasformer~\citep{Contrasformer}, GraphGPS~\citep{GraphGPS} and CAGT~\citep{CAGT}; (3) Mamba-based methods (MB): Mamba~\citep{mamba} and GraphMamba~\citep{GraphMamba}. As shown in Table~\ref{tab:Comparison-methods}, HLBG consistently outperforms existing methods on both ABIDE-I and REST-MDD datasets. Specifically, HLBG achieves improvements of 1.44\% in ACC and 1.76\% in SPE over the second-best method on ABIDE-I. On REST-MDD, it further improves ACC, SEN, and SPE by 1.90\%, 1.62\%, and 0.90\%, respectively. These results demonstrate that Hierarchical Brain Representation Learning captures hierarchical ROI$\rightarrow$community$\rightarrow$whole-brain relationships, while Graph-aware Mamba enhances topology-aware long-range dependency modeling, jointly improving brain network representation learning and classification performance.

\begin{table*}[t]
\centering
\footnotesize
\setlength{\tabcolsep}{8pt}
\renewcommand{\arraystretch}{1.25}
\caption{Ablation experiments of different components on the ABIDE-I dataset and REST-MDD dataset. \textbf{Bold} indicates the best result.}
\label{tab:ablation_study}

\begin{tabular}{c| c c c| c c c c}
\toprule
\toprule
\textbf{Dataset} & \textbf{GaMamba} & \textbf{SA} & \textbf{HBRL} 
& \textbf{ACC} & \textbf{AUC} & \textbf{SEN} & \textbf{SPE} \\
\midrule
\multirow{5}{*}{ABIDE-I}
 &  &  &  & 71.69\std{3.54} & 74.52\std{4.27} & 70.26\std{12.55} & \textbf{72.56\std{7.71}} \\
 & \checkmark &  &  & 73.91\std{2.25} & 77.74\std{3.61} & 75.99\std{6.35} & 71.36\std{4.91} \\
 & \checkmark & \checkmark &  & 74.20\std{2.97} & 77.68\std{3.37} & \textbf{80.53\std{9.30}} & 67.47\std{6.36} \\
 & \checkmark &  & \checkmark & 74.40\std{3.15} & 78.68\std{3.53} & 77.02\std{11.21} & 71.52\std{9.84} \\
 & \checkmark & \checkmark & \checkmark & \textbf{75.45\std{3.32}} & \textbf{78.97\std{4.43} }& 79.09\std{7.59} & 71.36\std{5.68} \\
\midrule

\multirow{5}{*}{REST-MDD}
 &  &  &  & 67.73\std{2.48} & 71.67\std{3.17} & 70.81\std{5.83} & 64.24\std{5.64} \\
 & \checkmark &  &  & 69.24\std{2.62} & 73.35\std{4.12} & 73.53\std{5.49} & 64.48\std{8.59} \\
 & \checkmark & \checkmark &  & 69.37\std{3.52} & 73.27\std{2.86} & \textbf{75.46\std{5.22}} & 62.00\std{7.19} \\
 & \checkmark &  & \checkmark & 68.66\std{2.26} & 72.60\std{2.65} & 74.19\std{5.86} & 62.18\std{4.99} \\
 & \checkmark & \checkmark & \checkmark & \textbf{70.13\std{2.27}} & \textbf{74.11\std{2.46}} & 73.54\std{6.69} & \textbf{65.78\std{6.99}}\\
\bottomrule

\end{tabular}
\end{table*}

\begin{figure*}
\centering
\includegraphics[scale=0.58]{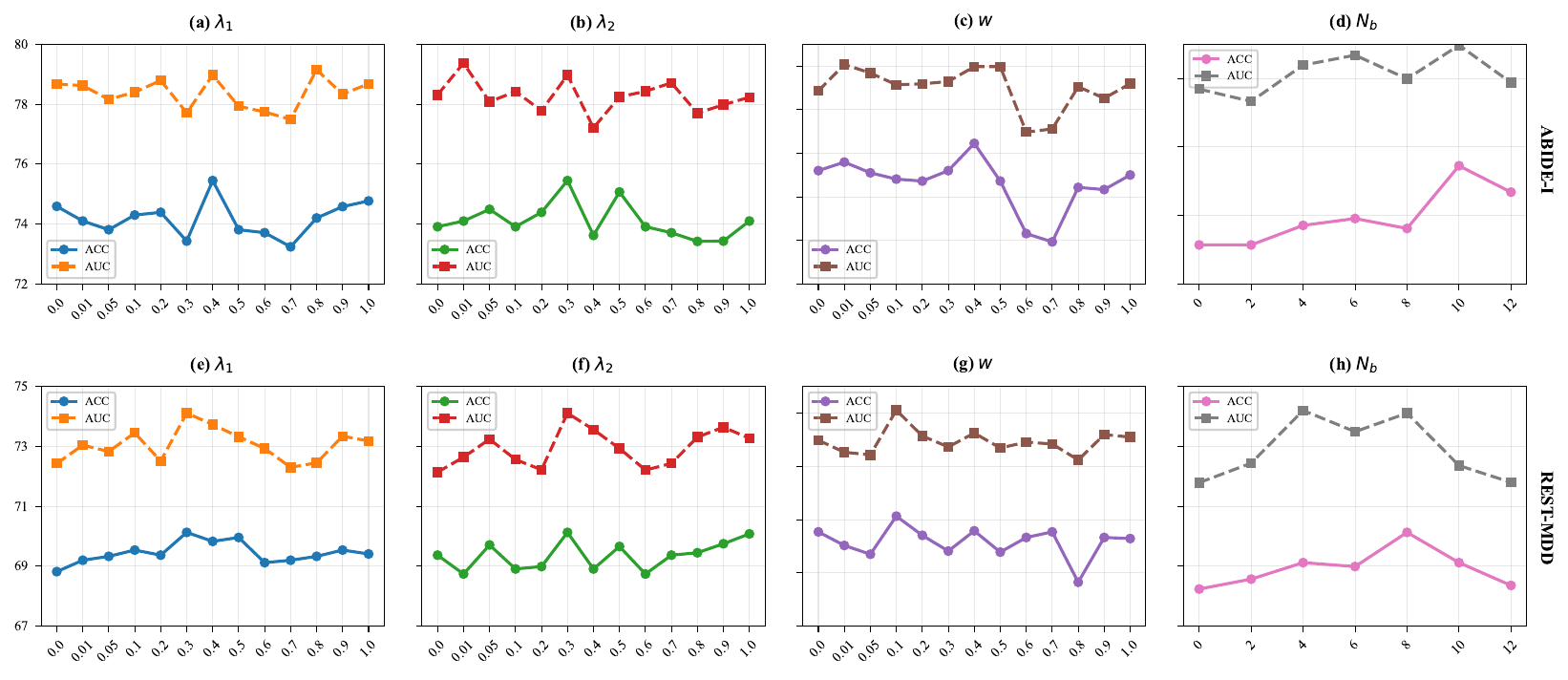}
\caption{Hyperparameter sensitivity analysis of \(\lambda_{\text{1}}\), \(\lambda_{\text{2}}\), \(w\) and \(N_b\). Panels (a)–(d) present results on the ABIDE dataset, while panels (e)–(h) show results on the REST-MDD dataset. }
\label{hyperparameters}
\end{figure*}
\subsection{Ablation Study}
\subsubsection{Ablation of Major Components}

To validate the effectiveness of each main component, we conduct ablation studies on ABIDE-I and REST-MDD. We evaluate three key modules: GaMamba, SA and HBRL.
Using only GaMamba brings clear improvements on both datasets. Specifically, ACC and AUC increase by 2.22\% and 3.22\% on ABIDE-I, and by 1.51\% and 1.68\% on REST-MDD, respectively. These results shows that GaMamba effectively incorporates graph structural prompts into Mamba and enhances topology-aware long-range dependency modeling. Adding SA further improves ACC and SEN in most cases. This indicates that SA can adaptively aggregate community-specific subgraph features and strengthen discriminative local patterns. When HBRL is combined with GaMamba, the model also achieves additional gains on ABIDE-I, suggesting that the two entailment losses provide useful hierarchical constraints in Lorentzian hyperbolic space. With all three components (GaMamba + SA + HBRL), the model achieves the best overall performance. Compared with the baseline, ACC improves by 3.76\% and 2.40\%, while AUC improves by 4.45\% and 2.44\% on ABIDE-I and REST-MDD, respectively. These results demonstrate that GaMamba, SA, and HBRL are complementary. GaMamba captures structure-aware long-range dependencies, SA fuses community-level representations, and HBRL guides the model to learn hierarchy-aware brain graph embeddings.

\subsubsection{Effectiveness of Graph-aware Mamba
module}
\begin{figure}
\includegraphics[scale=0.37]{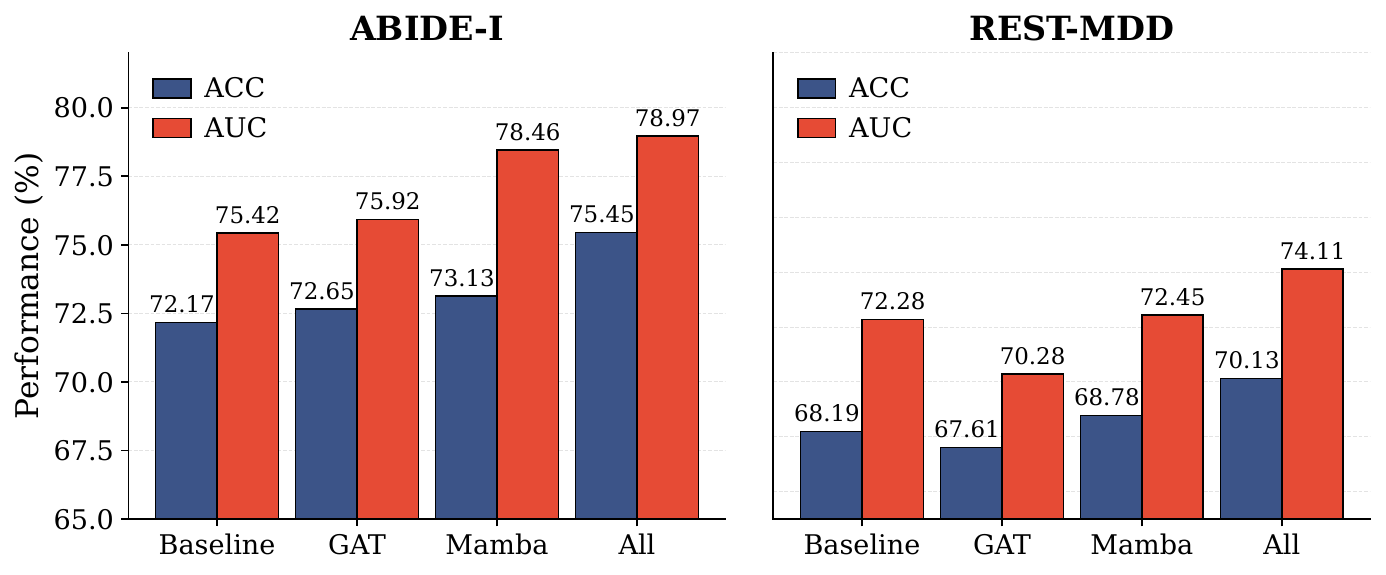}
\caption{Ablation analysis of the Graph-aware Mamba module.}
\label{GAM}
\end{figure}

To further evaluate the effectiveness of GaMamba, which integrates GAT and Mamba, we conduct an ablation analysis, and the results are shown in Fig.~\ref{GAM}. Removing GAT and retaining only Mamba decreases ACC by 2.32\% and 1.35\%, and AUC by 0.51\% and 1.66\% on ABIDE-I and REST-MDD, respectively. These results highlight the benefit of injecting topology-derived structural prompts into Mamba's selective state-space model. Nevertheless, Mamba alone remains competitive because its sequence modeling capability, combined with topology-aware ROI ordering, places functionally related regions close to each other and facilitates dependency modeling. Conversely, retaining only GAT reduces ACC by 2.80\% and 2.52\%, and AUC by 3.05\% and 3.83\% on ABIDE-I and REST-MDD, respectively. This confirms the importance of Mamba for capturing long-range interactions. Overall, GAT and Mamba provide complementary structural and sequential information, jointly improving brain network classification.

\subsection{Hyperparameter Analysis}

In this section, we analyze the sensitivity of four main hyperparameters:  the weighting coefficients \(w\) in Eq(\ref{12}), \(\lambda_1\), \(\lambda_2\) in Eq(\ref{13}), and the learnable prompt token size \(N_b\). We evaluate these hyperparameters on ABIDE-I and REST-MDD, with results presented in Fig.~\ref{hyperparameters}.  Specifically, \(\lambda_1\), \(\lambda_2\), and \(w\) are searched over ({0, 0.01, 0.05, 0.1, 0.2, 0.3, 0.4, 0.5, 0.6, 0.7, 0.8, 0.9, 1}), while \(N_b\) is selected from ({0, 2, 4, 6, 8, 10, 12}).
As shown in Fig.~\ref{hyperparameters}, HLBG maintains stable performance over a broad range of settings. For \(\lambda_1\) and \(\lambda_2\), moderate values yield better performance by effectively constraining ROI$\rightarrow$community$\rightarrow$whole-brain hierarchical relationships in Lorentzian hyperbolic space, whereas overly large values tend to overemphasize hierarchical consistency and weaken discriminative learning. Similarly, a moderate \(w\) improves global-local feature interaction, while a large value may introduce noise into the fused representation. For \(N_b\), the performance first improves and then declines, indicating that sufficient prompt tokens facilitate cross-subgraph information fusion, whereas excessive tokens may introduce redundancy and additional computational cost. The best performance is achieved with \(N_b=10\) on ABIDE-I and \(N_b=8\) on REST-MDD. Overall, these results demonstrate that HLBG is robust and not highly sensitive to hyperparameter selection.

\begin{figure*}
\centering
\includegraphics[scale=0.44]{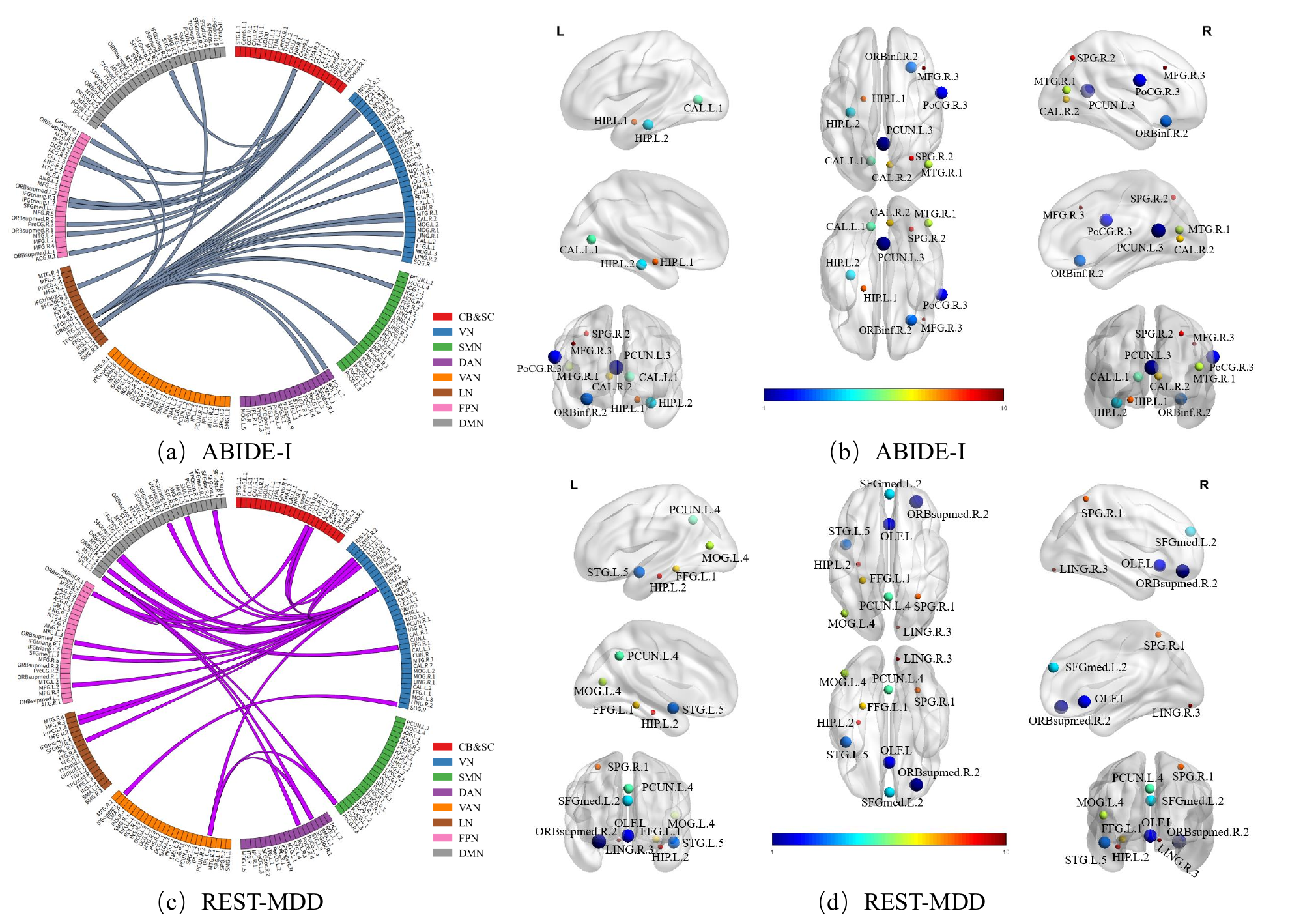}
\caption{Visualization of the brain connectivity networks and key brain regions on ABIDE-I and REST-MDD datasets. (a) and (c) visualize the top 30 most critical brain functional connections derived from the attention matrices $\mathbf{A}_{\text{ASD}}$ and $\mathbf{A}_{\text{MDD}}$, respectively. (b) and (d) visualize the top ten key ROIs identified from the attention matrices $\mathbf{A}_{\text{ASD}}$ and $\mathbf{A}_{\text{MDD}}$, respectively.}
\label{important_ROI}
\end{figure*}
\subsection{Biomarker Detection}

To evaluate the biomarker identification ability of HLBG, we visualize the important functional connections and brain regions highlighted by the attention matrices, as shown in Fig.~\ref{important_ROI}. Following \citep{CAGT}, we average the attention matrices of correctly classified ASD subjects on ABIDE-I and MDD subjects on REST-MDD, obtaining \(\mathbf{A}_{\mathrm{ASD}}\) and \(\mathbf{A}_{\mathrm{MDD}}\), respectively.

In Figs.~\ref{important_ROI} (a) and (b), the most important ASD-related connections are mainly distributed across DMN, FPN, and VN, indicating that HLBG captures distributed cross-network interactions rather than relying on a single functional subsystem~\citep{ilioska2023connectome}. Several high-contribution regions, such as precuneus (PCUN), middle frontal gyrus (MFG), and superior parietal gyrus (SPG), are highlighted, suggesting the involvement of self-related processing, executive control, and attention-related functions in ASD-related patterns~\citep{xu2020specific}.

As shown in Figs.~\ref{important_ROI} (c) and (d), the MDD-related connections involve multiple functional communities, including the DMN, FPN, VAN/DAN, LN, and VN, which is consistent with large-scale network dysfunction reported in depression~\citep{kaiser2015large,javaheripour2021altered}. HLBG further highlights several key regions, including the left olfactory cortex (OLF.L), medial orbital part of the superior frontal gyrus (ORBsupmed), medial superior frontal gyrus (SFGmed), PCUN, hippocampus (HIP), and fusiform gyrus (FFG). These regions are associated with altered self-referential, emotional, memory-related, and perceptual processing in MDD, in line with previous findings on precuneus- and orbitofrontal-related functional abnormalities~\citep{cheng2018functional}.

\section{Conclusion}
In this paper, we propose HLBG, a Hyperbolic Learning on Brain Graphs framework for brain network analysis. HLBG jointly learns ROI-, community-, and whole-brain-level representations to capture hierarchical relationships via the HBRL module. 
In addition, we design a new GaMamba for both 
community and whole-brain network feature extraction. 
It incorporates topology-derived structural prompts into Mamba for structure-aware long-range dependency modeling. The global and local representations are adaptively fused to facilitate brain graph representation and classification.  
Experiments on ABIDE-I and REST-MDD demonstrate the superiority of HLBG in disorder diagnosis and biomarker identification.

\printcredits

\bibliographystyle{cas-model2-names}

\bibliography{ref}

\begin{thebibliography}{53}
\expandafter\ifx\csname natexlab\endcsname\relax\def\natexlab#1{#1}\fi
\providecommand{\url}[1]{\texttt{#1}}
\providecommand{\href}[2]{#2}
\providecommand{\path}[1]{#1}
\providecommand{\DOIprefix}{doi:}
\providecommand{\ArXivprefix}{arXiv:}
\providecommand{\URLprefix}{URL: }
\providecommand{\Pubmedprefix}{pmid:}
\providecommand{\doi}[1]{\href{http://dx.doi.org/#1}{\path{#1}}}
\providecommand{\Pubmed}[1]{\href{pmid:#1}{\path{#1}}}
\providecommand{\bibinfo}[2]{#2}
\ifx\xfnm\relax \def\xfnm[#1]{\unskip,\space#1}\fi
\bibitem[{Baek et~al.(2025)Baek, Lee, Sim, Jeong and Kim}]{baek2025mnm}
\bibinfo{author}{Baek, S.}, \bibinfo{author}{Lee, J.}, \bibinfo{author}{Sim, J.}, \bibinfo{author}{Jeong, M.}, \bibinfo{author}{Kim, W.H.}, \bibinfo{year}{2025}.
\newblock \bibinfo{title}{Mnm: Multi-level neuroimaging meta-analysis with hyperbolic brain-text representations}, in: \bibinfo{booktitle}{International Conference on Medical Image Computing and Computer-Assisted Intervention}, \bibinfo{organization}{Springer}. pp. \bibinfo{pages}{397--406}.
\bibitem[{Baker et~al.(2024)Baker, Suárez-Méndez, Smith, Marsh, Funke, Mosher, Maestú, Xu and Pantazis}]{hyperbolic_JBHI}
\bibinfo{author}{Baker, C.}, \bibinfo{author}{Suárez-Méndez, I.}, \bibinfo{author}{Smith, G.}, \bibinfo{author}{Marsh, E.B.}, \bibinfo{author}{Funke, M.}, \bibinfo{author}{Mosher, J.C.}, \bibinfo{author}{Maestú, F.}, \bibinfo{author}{Xu, M.}, \bibinfo{author}{Pantazis, D.}, \bibinfo{year}{2024}.
\newblock \bibinfo{title}{Hyperbolic graph embedding of meg brain networks to study brain alterations in individuals with subjective cognitive decline}.
\newblock \bibinfo{journal}{IEEE Journal of Biomedical and Health Informatics} \bibinfo{volume}{28}, \bibinfo{pages}{7357--7368}.
\newblock \DOIprefix\doi{10.1109/JBHI.2024.3416890}.
\bibitem[{Bannadabhavi et~al.(2023)Bannadabhavi, Lee, Deng, Ying and Li}]{Com-BrainTF}
\bibinfo{author}{Bannadabhavi, A.}, \bibinfo{author}{Lee, S.}, \bibinfo{author}{Deng, W.}, \bibinfo{author}{Ying, R.}, \bibinfo{author}{Li, X.}, \bibinfo{year}{2023}.
\newblock \bibinfo{title}{Community-aware transformer for autism prediction in fmri connectome}, in: \bibinfo{booktitle}{International conference on medical image computing and computer-assisted intervention}, \bibinfo{organization}{Springer}. pp. \bibinfo{pages}{287--297}.
\bibitem[{Behrouz and Hashemi(2024)}]{graphmambakdd}
\bibinfo{author}{Behrouz, A.}, \bibinfo{author}{Hashemi, F.}, \bibinfo{year}{2024}.
\newblock \bibinfo{title}{Graph mamba: Towards learning on graphs with state space models}, in: \bibinfo{booktitle}{Proceedings of the 30th ACM SIGKDD Conference on Knowledge Discovery and Data Mining}, \bibinfo{publisher}{Association for Computing Machinery}, \bibinfo{address}{New York, NY, USA}. p. \bibinfo{pages}{119–130}.
\bibitem[{Benesty et~al.(2009)Benesty, Chen, Huang and Cohen}]{PCC}
\bibinfo{author}{Benesty, J.}, \bibinfo{author}{Chen, J.}, \bibinfo{author}{Huang, Y.}, \bibinfo{author}{Cohen, I.}, \bibinfo{year}{2009}.
\newblock \bibinfo{title}{Pearson correlation coefficient}, in: \bibinfo{booktitle}{Noise reduction in speech processing}. \bibinfo{publisher}{Springer}, pp. \bibinfo{pages}{1--4}.
\bibitem[{Cannon et~al.(1997)Cannon, Floyd, Kenyon, Parry et~al.}]{1997hyperbolic}
\bibinfo{author}{Cannon, J.W.}, \bibinfo{author}{Floyd, W.J.}, \bibinfo{author}{Kenyon, R.}, \bibinfo{author}{Parry, W.R.}, et~al., \bibinfo{year}{1997}.
\newblock \bibinfo{title}{Hyperbolic geometry}.
\newblock \bibinfo{journal}{Flavors of geometry} \bibinfo{volume}{31}, \bibinfo{pages}{2}.
\bibitem[{Chen et~al.(2025)Chen, Wang, Cao and Wu}]{subgraph_fusion}
\bibinfo{author}{Chen, J.}, \bibinfo{author}{Wang, J.}, \bibinfo{author}{Cao, Z.}, \bibinfo{author}{Wu, Y.}, \bibinfo{year}{2025}.
\newblock \bibinfo{title}{Neural multi-objective combinatorial optimization via graph-image multimodal fusion}, in: \bibinfo{editor}{Yue, Y.}, \bibinfo{editor}{Garg, A.}, \bibinfo{editor}{Peng, N.}, \bibinfo{editor}{Sha, F.}, \bibinfo{editor}{Yu, R.} (Eds.), \bibinfo{booktitle}{International Conference on Learning Representations}, pp. \bibinfo{pages}{33964--33983}.
\bibitem[{Chen et~al.(2022)Chen, Lu, Li, Li, Wang, Castellanos, Cao, Chen, Chen, Cheng et~al.}]{MDD}
\bibinfo{author}{Chen, X.}, \bibinfo{author}{Lu, B.}, \bibinfo{author}{Li, H.X.}, \bibinfo{author}{Li, X.Y.}, \bibinfo{author}{Wang, Y.W.}, \bibinfo{author}{Castellanos, F.X.}, \bibinfo{author}{Cao, L.P.}, \bibinfo{author}{Chen, N.X.}, \bibinfo{author}{Chen, W.}, \bibinfo{author}{Cheng, Y.Q.}, et~al., \bibinfo{year}{2022}.
\newblock \bibinfo{title}{The direct consortium and the rest-meta-mdd project: towards neuroimaging biomarkers of major depressive disorder}.
\newblock \bibinfo{journal}{Psychoradiology} \bibinfo{volume}{2}, \bibinfo{pages}{32--42}.
\bibitem[{Cheng et~al.(2018)Cheng, Rolls, Qiu, Yang, Ruan, Wei, Zhao, Meng, Xie and Feng}]{cheng2018functional}
\bibinfo{author}{Cheng, W.}, \bibinfo{author}{Rolls, E.T.}, \bibinfo{author}{Qiu, J.}, \bibinfo{author}{Yang, D.}, \bibinfo{author}{Ruan, H.}, \bibinfo{author}{Wei, D.}, \bibinfo{author}{Zhao, L.}, \bibinfo{author}{Meng, J.}, \bibinfo{author}{Xie, P.}, \bibinfo{author}{Feng, J.}, \bibinfo{year}{2018}.
\newblock \bibinfo{title}{Functional connectivity of the precuneus in unmedicated patients with depression}.
\newblock \bibinfo{journal}{Biological psychiatry: cognitive neuroscience and neuroimaging} \bibinfo{volume}{3}, \bibinfo{pages}{1040--1049}.
\bibitem[{Craddock et~al.(2013)Craddock, Benhajali, Chu, Chouinard, Evans, Jakab, Khundrakpam, Lewis, Li, Milham et~al.}]{ABIDE}
\bibinfo{author}{Craddock, C.}, \bibinfo{author}{Benhajali, Y.}, \bibinfo{author}{Chu, C.}, \bibinfo{author}{Chouinard, F.}, \bibinfo{author}{Evans, A.}, \bibinfo{author}{Jakab, A.}, \bibinfo{author}{Khundrakpam, B.S.}, \bibinfo{author}{Lewis, J.D.}, \bibinfo{author}{Li, Q.}, \bibinfo{author}{Milham, M.}, et~al., \bibinfo{year}{2013}.
\newblock \bibinfo{title}{The neuro bureau preprocessing initiative: open sharing of preprocessed neuroimaging data and derivatives}.
\newblock \bibinfo{journal}{Frontiers in Neuroinformatics} \bibinfo{volume}{7}, \bibinfo{pages}{5}.
\bibitem[{Craddock et~al.(2012)Craddock, James, Holtzheimer~III, Hu and Mayberg}]{CC200}
\bibinfo{author}{Craddock, R.C.}, \bibinfo{author}{James, G.A.}, \bibinfo{author}{Holtzheimer~III, P.E.}, \bibinfo{author}{Hu, X.P.}, \bibinfo{author}{Mayberg, H.S.}, \bibinfo{year}{2012}.
\newblock \bibinfo{title}{A whole brain fmri atlas generated via spatially constrained spectral clustering}.
\newblock \bibinfo{journal}{Human brain mapping} \bibinfo{volume}{33}, \bibinfo{pages}{1914--1928}.
\bibitem[{Cui et~al.(2022a)Cui, Dai, Zhu, Kan, Gu, Lukemire, Zhan, He, Guo and Yang}]{braingb}
\bibinfo{author}{Cui, H.}, \bibinfo{author}{Dai, W.}, \bibinfo{author}{Zhu, Y.}, \bibinfo{author}{Kan, X.}, \bibinfo{author}{Gu, A.A.C.}, \bibinfo{author}{Lukemire, J.}, \bibinfo{author}{Zhan, L.}, \bibinfo{author}{He, L.}, \bibinfo{author}{Guo, Y.}, \bibinfo{author}{Yang, C.}, \bibinfo{year}{2022}a.
\newblock \bibinfo{title}{Braingb: a benchmark for brain network analysis with graph neural networks}.
\newblock \bibinfo{journal}{IEEE transactions on medical imaging} \bibinfo{volume}{42}, \bibinfo{pages}{493--506}.
\bibitem[{Cui et~al.(2022b)Cui, Dai, Zhu, Li, He and Yang}]{IBGNN}
\bibinfo{author}{Cui, H.}, \bibinfo{author}{Dai, W.}, \bibinfo{author}{Zhu, Y.}, \bibinfo{author}{Li, X.}, \bibinfo{author}{He, L.}, \bibinfo{author}{Yang, C.}, \bibinfo{year}{2022}b.
\newblock \bibinfo{title}{Interpretable graph neural networks for connectome-based brain disorder analysis}, in: \bibinfo{booktitle}{International conference on medical image computing and computer-assisted intervention}, \bibinfo{organization}{Springer}. pp. \bibinfo{pages}{375--385}.
\bibitem[{Dao and Gu(2024)}]{mamba2}
\bibinfo{author}{Dao, T.}, \bibinfo{author}{Gu, A.}, \bibinfo{year}{2024}.
\newblock \bibinfo{title}{Transformers are {SSM}s: Generalized models and efficient algorithms through structured state space duality}, in: \bibinfo{editor}{Salakhutdinov, R.}, \bibinfo{editor}{Kolter, Z.}, \bibinfo{editor}{Heller, K.}, \bibinfo{editor}{Weller, A.}, \bibinfo{editor}{Oliver, N.}, \bibinfo{editor}{Scarlett, J.}, \bibinfo{editor}{Berkenkamp, F.} (Eds.), \bibinfo{booktitle}{Proceedings of the 41st International Conference on Machine Learning}, \bibinfo{publisher}{PMLR}. pp. \bibinfo{pages}{10041--10071}.
\bibitem[{Desai et~al.(2023)Desai, Nickel, Rajpurohit, Johnson and Vedantam}]{2023hyperbolic}
\bibinfo{author}{Desai, K.}, \bibinfo{author}{Nickel, M.}, \bibinfo{author}{Rajpurohit, T.}, \bibinfo{author}{Johnson, J.}, \bibinfo{author}{Vedantam, S.R.}, \bibinfo{year}{2023}.
\newblock \bibinfo{title}{Hyperbolic image-text representations}, in: \bibinfo{booktitle}{International Conference on Machine Learning}, \bibinfo{organization}{PMLR}. pp. \bibinfo{pages}{7694--7731}.
\bibitem[{Ganea et~al.(2018)Ganea, B{\'e}cigneul and Hofmann}]{ganea2018hyperbolic}
\bibinfo{author}{Ganea, O.}, \bibinfo{author}{B{\'e}cigneul, G.}, \bibinfo{author}{Hofmann, T.}, \bibinfo{year}{2018}.
\newblock \bibinfo{title}{Hyperbolic entailment cones for learning hierarchical embeddings}, in: \bibinfo{booktitle}{International conference on machine learning}, \bibinfo{organization}{PMLR}. pp. \bibinfo{pages}{1646--1655}.
\bibitem[{Gu and Dao(2024)}]{mamba}
\bibinfo{author}{Gu, A.}, \bibinfo{author}{Dao, T.}, \bibinfo{year}{2024}.
\newblock \bibinfo{title}{Mamba: Linear-time sequence modeling with selective state spaces}, in: \bibinfo{booktitle}{First conference on language modeling}.
\bibitem[{Guo et~al.(2025)Guo, Guo, Zha, Zhang, Li, Dai, Xia and Li}]{guo2025mambairv2}
\bibinfo{author}{Guo, H.}, \bibinfo{author}{Guo, Y.}, \bibinfo{author}{Zha, Y.}, \bibinfo{author}{Zhang, Y.}, \bibinfo{author}{Li, W.}, \bibinfo{author}{Dai, T.}, \bibinfo{author}{Xia, S.T.}, \bibinfo{author}{Li, Y.}, \bibinfo{year}{2025}.
\newblock \bibinfo{title}{Mambairv2: Attentive state space restoration}, in: \bibinfo{booktitle}{Proceedings of the Computer Vision and Pattern Recognition Conference}, pp. \bibinfo{pages}{28124--28133}.
\bibitem[{Hu et~al.(2020)Hu, Liu, Gomes, Zitnik, Liang, Pande and Leskovec}]{GINE}
\bibinfo{author}{Hu, W.}, \bibinfo{author}{Liu, B.}, \bibinfo{author}{Gomes, J.}, \bibinfo{author}{Zitnik, M.}, \bibinfo{author}{Liang, P.}, \bibinfo{author}{Pande, V.}, \bibinfo{author}{Leskovec, J.}, \bibinfo{year}{2020}.
\newblock \bibinfo{title}{Strategies for pre-training graph neural networks}, in: \bibinfo{booktitle}{International Conference on Learning Representations}.
\bibitem[{Ilioska et~al.(2023)Ilioska, Oldehinkel, Llera, Chopra, Looden, Chauvin, Van~Rooij, Floris, Tillmann, Moessnang et~al.}]{ilioska2023connectome}
\bibinfo{author}{Ilioska, I.}, \bibinfo{author}{Oldehinkel, M.}, \bibinfo{author}{Llera, A.}, \bibinfo{author}{Chopra, S.}, \bibinfo{author}{Looden, T.}, \bibinfo{author}{Chauvin, R.}, \bibinfo{author}{Van~Rooij, D.}, \bibinfo{author}{Floris, D.L.}, \bibinfo{author}{Tillmann, J.}, \bibinfo{author}{Moessnang, C.}, et~al., \bibinfo{year}{2023}.
\newblock \bibinfo{title}{Connectome-wide mega-analysis reveals robust patterns of atypical functional connectivity in autism}.
\newblock \bibinfo{journal}{Biological psychiatry} \bibinfo{volume}{94}, \bibinfo{pages}{29--39}.
\bibitem[{Insel and Cuthbert(2015)}]{background1}
\bibinfo{author}{Insel, T.R.}, \bibinfo{author}{Cuthbert, B.N.}, \bibinfo{year}{2015}.
\newblock \bibinfo{title}{Brain disorders? precisely}.
\newblock \bibinfo{journal}{Science} \bibinfo{volume}{348}, \bibinfo{pages}{499--500}.
\bibitem[{Javaheripour et~al.(2021)Javaheripour, Li, Chand, Krug, Kircher, Dannlowski, Nenadi{\'c}, Hamilton, Sacchet, Gotlib et~al.}]{javaheripour2021altered}
\bibinfo{author}{Javaheripour, N.}, \bibinfo{author}{Li, M.}, \bibinfo{author}{Chand, T.}, \bibinfo{author}{Krug, A.}, \bibinfo{author}{Kircher, T.}, \bibinfo{author}{Dannlowski, U.}, \bibinfo{author}{Nenadi{\'c}, I.}, \bibinfo{author}{Hamilton, J.P.}, \bibinfo{author}{Sacchet, M.D.}, \bibinfo{author}{Gotlib, I.H.}, et~al., \bibinfo{year}{2021}.
\newblock \bibinfo{title}{Altered resting-state functional connectome in major depressive disorder: a mega-analysis from the psymri consortium}.
\newblock \bibinfo{journal}{Translational psychiatry} \bibinfo{volume}{11}, \bibinfo{pages}{511}.
\bibitem[{Jia et~al.(2026)Jia, Liu, Yang, Sun, Qin, Wang and Peng}]{Brain-HGCN}
\bibinfo{author}{Jia, J.}, \bibinfo{author}{Liu, Y.}, \bibinfo{author}{Yang, C.}, \bibinfo{author}{Sun, Y.}, \bibinfo{author}{Qin, F.}, \bibinfo{author}{Wang, C.}, \bibinfo{author}{Peng, Y.}, \bibinfo{year}{2026}.
\newblock \bibinfo{title}{Brain-hgcn: A hyperbolic graph convolutional network for brain functional network analysis}, in: \bibinfo{booktitle}{IEEE International Conference on Acoustics, Speech and Signal Processing (ICASSP)}, \bibinfo{organization}{IEEE}. pp. \bibinfo{pages}{6391--6395}.
\bibitem[{Kaiser et~al.(2015)Kaiser, Andrews-Hanna, Wager and Pizzagalli}]{kaiser2015large}
\bibinfo{author}{Kaiser, R.H.}, \bibinfo{author}{Andrews-Hanna, J.R.}, \bibinfo{author}{Wager, T.D.}, \bibinfo{author}{Pizzagalli, D.A.}, \bibinfo{year}{2015}.
\newblock \bibinfo{title}{Large-scale network dysfunction in major depressive disorder: a meta-analysis of resting-state functional connectivity}.
\newblock \bibinfo{journal}{JAMA psychiatry} \bibinfo{volume}{72}, \bibinfo{pages}{603--611}.
\bibitem[{Kan et~al.(2022)Kan, Dai, Cui, Zhang, Guo and Yang}]{BrainNetTF}
\bibinfo{author}{Kan, X.}, \bibinfo{author}{Dai, W.}, \bibinfo{author}{Cui, H.}, \bibinfo{author}{Zhang, Z.}, \bibinfo{author}{Guo, Y.}, \bibinfo{author}{Yang, C.}, \bibinfo{year}{2022}.
\newblock \bibinfo{title}{Brain network transformer}.
\newblock \bibinfo{journal}{Advances in Neural Information Processing Systems} \bibinfo{volume}{35}, \bibinfo{pages}{25586--25599}.
\bibitem[{Lawrence et~al.(2021)Lawrence, Bridgeford, Myers, Arvapalli, Ramachandran, Pisner, Frank, Lemmer, Nikolaidis and Vogelstein}]{8subgraph2}
\bibinfo{author}{Lawrence, R.M.}, \bibinfo{author}{Bridgeford, E.W.}, \bibinfo{author}{Myers, P.E.}, \bibinfo{author}{Arvapalli, G.C.}, \bibinfo{author}{Ramachandran, S.C.}, \bibinfo{author}{Pisner, D.A.}, \bibinfo{author}{Frank, P.F.}, \bibinfo{author}{Lemmer, A.D.}, \bibinfo{author}{Nikolaidis, A.}, \bibinfo{author}{Vogelstein, J.T.}, \bibinfo{year}{2021}.
\newblock \bibinfo{title}{Standardizing human brain parcellations}.
\newblock \bibinfo{journal}{Scientific data} \bibinfo{volume}{8}, \bibinfo{pages}{78}.
\bibitem[{Li et~al.(2021a)Li, Zhou, Dvornek, Zhang, Gao, Zhuang, Scheinost, Staib, Ventola and Duncan}]{BrainGNN}
\bibinfo{author}{Li, X.}, \bibinfo{author}{Zhou, Y.}, \bibinfo{author}{Dvornek, N.}, \bibinfo{author}{Zhang, M.}, \bibinfo{author}{Gao, S.}, \bibinfo{author}{Zhuang, J.}, \bibinfo{author}{Scheinost, D.}, \bibinfo{author}{Staib, L.H.}, \bibinfo{author}{Ventola, P.}, \bibinfo{author}{Duncan, J.S.}, \bibinfo{year}{2021}a.
\newblock \bibinfo{title}{Braingnn: Interpretable brain graph neural network for fmri analysis}.
\newblock \bibinfo{journal}{Medical Image Analysis} \bibinfo{volume}{74}, \bibinfo{pages}{102233}.
\bibitem[{Li et~al.(2021b)Li, Liu, Jiang, Liu and Lei}]{background2}
\bibinfo{author}{Li, Y.}, \bibinfo{author}{Liu, J.}, \bibinfo{author}{Jiang, Y.}, \bibinfo{author}{Liu, Y.}, \bibinfo{author}{Lei, B.}, \bibinfo{year}{2021}b.
\newblock \bibinfo{title}{Virtual adversarial training-based deep feature aggregation network from dynamic effective connectivity for mci identification}.
\newblock \bibinfo{journal}{IEEE transactions on medical imaging} \bibinfo{volume}{41}, \bibinfo{pages}{237--251}.
\bibitem[{Luo et~al.(2024)Luo, Wu, Yang, Xue, Beheshti, Sheng, McAlpine, Sowman, Giral and Yu}]{survey}
\bibinfo{author}{Luo, X.}, \bibinfo{author}{Wu, J.}, \bibinfo{author}{Yang, J.}, \bibinfo{author}{Xue, S.}, \bibinfo{author}{Beheshti, A.}, \bibinfo{author}{Sheng, Q.Z.}, \bibinfo{author}{McAlpine, D.}, \bibinfo{author}{Sowman, P.}, \bibinfo{author}{Giral, A.}, \bibinfo{author}{Yu, P.S.}, \bibinfo{year}{2024}.
\newblock \bibinfo{title}{Graph neural networks for brain graph learning: A survey}, in: \bibinfo{editor}{Larson, K.} (Ed.), \bibinfo{booktitle}{Proceedings of the Thirty-Third International Joint Conference on Artificial Intelligence, {IJCAI-24}}, \bibinfo{publisher}{International Joint Conferences on Artificial Intelligence Organization}. pp. \bibinfo{pages}{8170--8178}.
\bibitem[{Ma et~al.(2023)Ma, Cui, Liu, Guo, Chen and Li}]{ma2023multi}
\bibinfo{author}{Ma, Y.}, \bibinfo{author}{Cui, W.}, \bibinfo{author}{Liu, J.}, \bibinfo{author}{Guo, Y.}, \bibinfo{author}{Chen, H.}, \bibinfo{author}{Li, Y.}, \bibinfo{year}{2023}.
\newblock \bibinfo{title}{A multi-graph cross-attention-based region-aware feature fusion network using multi-template for brain disorder diagnosis}.
\newblock \bibinfo{journal}{IEEE Transactions on Medical Imaging} \bibinfo{volume}{43}, \bibinfo{pages}{1045--1059}.
\bibitem[{Pal et~al.(2025)Pal, van Spengler, di~Melendugno, Flaborea, Galasso and Mettes}]{2025CLIPhyperbolic}
\bibinfo{author}{Pal, A.}, \bibinfo{author}{van Spengler, M.}, \bibinfo{author}{di~Melendugno, G.M.D.}, \bibinfo{author}{Flaborea, A.}, \bibinfo{author}{Galasso, F.}, \bibinfo{author}{Mettes, P.}, \bibinfo{year}{2025}.
\newblock \bibinfo{title}{Compositional entailment learning for hyperbolic vision-language models}, in: \bibinfo{editor}{Yue, Y.}, \bibinfo{editor}{Garg, A.}, \bibinfo{editor}{Peng, N.}, \bibinfo{editor}{Sha, F.}, \bibinfo{editor}{Yu, R.} (Eds.), \bibinfo{booktitle}{International Conference on Learning Representations}, pp. \bibinfo{pages}{87371--87399}.
\bibitem[{Pei et~al.(2025)Pei, Ma, Lv, Zhang and Guan}]{CAGT}
\bibinfo{author}{Pei, S.}, \bibinfo{author}{Ma, J.}, \bibinfo{author}{Lv, Z.}, \bibinfo{author}{Zhang, C.}, \bibinfo{author}{Guan, J.}, \bibinfo{year}{2025}.
\newblock \bibinfo{title}{Community-aware graph transformer for brain disorder identification}, in: \bibinfo{editor}{Kwok, J.} (Ed.), \bibinfo{booktitle}{Proceedings of the Thirty-Fourth International Joint Conference on Artificial Intelligence, {IJCAI-25}}, \bibinfo{publisher}{International Joint Conferences on Artificial Intelligence Organization}. pp. \bibinfo{pages}{4191--4199}.
\bibitem[{Peng et~al.(2025)Peng, Huang, Dong, Yu, Xia, Zhang and Jin}]{BioBGT}
\bibinfo{author}{Peng, C.}, \bibinfo{author}{Huang, Y.}, \bibinfo{author}{Dong, Q.}, \bibinfo{author}{Yu, S.}, \bibinfo{author}{Xia, F.}, \bibinfo{author}{Zhang, C.}, \bibinfo{author}{Jin, Y.}, \bibinfo{year}{2025}.
\newblock \bibinfo{title}{Biologically plausible brain graph transformer}, in: \bibinfo{editor}{Yue, Y.}, \bibinfo{editor}{Garg, A.}, \bibinfo{editor}{Peng, N.}, \bibinfo{editor}{Sha, F.}, \bibinfo{editor}{Yu, R.} (Eds.), \bibinfo{booktitle}{International Conference on Learning Representations}, pp. \bibinfo{pages}{52283--52309}.
\bibitem[{Peng et~al.(2024)Peng, He, Jiang, Wang and Yuan}]{GBT}
\bibinfo{author}{Peng, Z.}, \bibinfo{author}{He, Z.}, \bibinfo{author}{Jiang, Y.}, \bibinfo{author}{Wang, P.}, \bibinfo{author}{Yuan, Y.}, \bibinfo{year}{2024}.
\newblock \bibinfo{title}{Gbt: Geometric-oriented brain transformer for autism diagnosis}, in: \bibinfo{booktitle}{International Conference on Medical Image Computing and Computer-Assisted Intervention}, \bibinfo{organization}{Springer}. pp. \bibinfo{pages}{142--152}.
\bibitem[{Qiu et~al.(2025)Qiu, Sun, Shi, Duan, Wang and Ma}]{QIU2025103664}
\bibinfo{author}{Qiu, X.}, \bibinfo{author}{Sun, Y.}, \bibinfo{author}{Shi, Y.}, \bibinfo{author}{Duan, X.}, \bibinfo{author}{Wang, F.}, \bibinfo{author}{Ma, J.}, \bibinfo{year}{2025}.
\newblock \bibinfo{title}{Metaexplainer: Revisit domain generalization of functional connectome analyses from the perspective of explainability}.
\newblock \bibinfo{journal}{Medical Image Analysis} \bibinfo{volume}{105}, \bibinfo{pages}{103664}.
\bibitem[{Ramp{\'a}{\v{s}}ek et~al.(2022)Ramp{\'a}{\v{s}}ek, Galkin, Dwivedi, Luu, Wolf and Beaini}]{GraphGPS}
\bibinfo{author}{Ramp{\'a}{\v{s}}ek, L.}, \bibinfo{author}{Galkin, M.}, \bibinfo{author}{Dwivedi, V.P.}, \bibinfo{author}{Luu, A.T.}, \bibinfo{author}{Wolf, G.}, \bibinfo{author}{Beaini, D.}, \bibinfo{year}{2022}.
\newblock \bibinfo{title}{Recipe for a general, powerful, scalable graph transformer}.
\newblock \bibinfo{journal}{Advances in Neural Information Processing Systems} \bibinfo{volume}{35}, \bibinfo{pages}{14501--14515}.
\bibitem[{Thapaliya et~al.(2025)Thapaliya, Akbas, Chen, Sapkota, Ray, Suresh, Calhoun and Liu}]{THAPALIYA2025103433}
\bibinfo{author}{Thapaliya, B.}, \bibinfo{author}{Akbas, E.}, \bibinfo{author}{Chen, J.}, \bibinfo{author}{Sapkota, R.}, \bibinfo{author}{Ray, B.}, \bibinfo{author}{Suresh, P.}, \bibinfo{author}{Calhoun, V.D.}, \bibinfo{author}{Liu, J.}, \bibinfo{year}{2025}.
\newblock \bibinfo{title}{Brain networks and intelligence: A graph neural network based approach to resting state fmri data}.
\newblock \bibinfo{journal}{Medical Image Analysis} \bibinfo{volume}{101}, \bibinfo{pages}{103433}.
\bibitem[{Vaswani et~al.(2017)Vaswani, Shazeer, Parmar, Uszkoreit, Jones, Gomez, Kaiser and Polosukhin}]{transformer}
\bibinfo{author}{Vaswani, A.}, \bibinfo{author}{Shazeer, N.}, \bibinfo{author}{Parmar, N.}, \bibinfo{author}{Uszkoreit, J.}, \bibinfo{author}{Jones, L.}, \bibinfo{author}{Gomez, A.N.}, \bibinfo{author}{Kaiser, {\L}.}, \bibinfo{author}{Polosukhin, I.}, \bibinfo{year}{2017}.
\newblock \bibinfo{title}{Attention is all you need}.
\newblock \bibinfo{journal}{Advances in neural information processing systems} \bibinfo{volume}{30}.
\bibitem[{Velickovic et~al.(2018)Velickovic, Cucurull, Casanova, Romero, Li{\`{o}} and Bengio}]{GAT}
\bibinfo{author}{Velickovic, P.}, \bibinfo{author}{Cucurull, G.}, \bibinfo{author}{Casanova, A.}, \bibinfo{author}{Romero, A.}, \bibinfo{author}{Li{\`{o}}, P.}, \bibinfo{author}{Bengio, Y.}, \bibinfo{year}{2018}.
\newblock \bibinfo{title}{Graph attention networks}, in: \bibinfo{booktitle}{6th International Conference on Learning Representations, {ICLR} 2018, Vancouver, BC, Canada, April 30 - May 3, 2018, Conference Track Proceedings}.
\bibitem[{Wang et~al.(2026)Wang, Liang, Ye, Yan, Liu and Yan}]{Trifocal_Transformer}
\bibinfo{author}{Wang, B.}, \bibinfo{author}{Liang, J.}, \bibinfo{author}{Ye, C.}, \bibinfo{author}{Yan, T.}, \bibinfo{author}{Liu, M.}, \bibinfo{author}{Yan, T.}, \bibinfo{year}{2026}.
\newblock \bibinfo{title}{Trifocal transformer: Connection-mask-residual focused attention network for brain disease diagnosis}.
\newblock \bibinfo{journal}{IEEE Journal of Biomedical and Health Informatics} \bibinfo{volume}{30}, \bibinfo{pages}{596--608}.
\newblock \DOIprefix\doi{10.1109/JBHI.2025.3624242}.
\bibitem[{Wang et~al.(2024)Wang, Tsepa, Ma and Wang}]{GraphMamba}
\bibinfo{author}{Wang, C.}, \bibinfo{author}{Tsepa, O.}, \bibinfo{author}{Ma, J.}, \bibinfo{author}{Wang, B.}, \bibinfo{year}{2024}.
\newblock \bibinfo{title}{Graph-mamba: Towards long-range graph sequence modeling with selective state spaces}.
\newblock \bibinfo{journal}{arXiv preprint arXiv:2402.00789} .
\bibitem[{Wing and Gould(1979)}]{background0}
\bibinfo{author}{Wing, L.}, \bibinfo{author}{Gould, J.}, \bibinfo{year}{1979}.
\newblock \bibinfo{title}{Severe impairments of social interaction and associated abnormalities in children: Epidemiology and classification}.
\newblock \bibinfo{journal}{Journal of autism and developmental disorders} \bibinfo{volume}{9}, \bibinfo{pages}{11--29}.
\bibitem[{Xiao et~al.(2025)Xiao, Li, ZHANG, Meng and Zhang}]{ICLR2025_Structure-Aware_mamba}
\bibinfo{author}{Xiao, C.}, \bibinfo{author}{Li, M.}, \bibinfo{author}{ZHANG, z.}, \bibinfo{author}{Meng, D.}, \bibinfo{author}{Zhang, L.}, \bibinfo{year}{2025}.
\newblock \bibinfo{title}{Spatial-mamba: Effective visual state space models via structure-aware state fusion}, in: \bibinfo{editor}{Yue, Y.}, \bibinfo{editor}{Garg, A.}, \bibinfo{editor}{Peng, N.}, \bibinfo{editor}{Sha, F.}, \bibinfo{editor}{Yu, R.} (Eds.), \bibinfo{booktitle}{International Conference on Learning Representations}, pp. \bibinfo{pages}{73777--73795}.
\bibitem[{Xu et~al.(2024)Xu, He, Lan, Bian, Li, Li, Ke and Qiao}]{Contrasformer}
\bibinfo{author}{Xu, J.}, \bibinfo{author}{He, K.}, \bibinfo{author}{Lan, M.}, \bibinfo{author}{Bian, Q.}, \bibinfo{author}{Li, W.}, \bibinfo{author}{Li, T.}, \bibinfo{author}{Ke, Y.}, \bibinfo{author}{Qiao, M.}, \bibinfo{year}{2024}.
\newblock \bibinfo{title}{Contrasformer: a brain network contrastive transformer for neurodegenerative condition identification}, in: \bibinfo{booktitle}{Proceedings of the 33rd ACM International Conference on Information and Knowledge Management}, pp. \bibinfo{pages}{2671--2681}.
\bibitem[{Xu et~al.(2020)Xu, Wang, Xu, Li, Chen, Chen, Gao, Wang and Hu}]{xu2020specific}
\bibinfo{author}{Xu, J.}, \bibinfo{author}{Wang, C.}, \bibinfo{author}{Xu, Z.}, \bibinfo{author}{Li, T.}, \bibinfo{author}{Chen, F.}, \bibinfo{author}{Chen, K.}, \bibinfo{author}{Gao, J.}, \bibinfo{author}{Wang, J.}, \bibinfo{author}{Hu, Q.}, \bibinfo{year}{2020}.
\newblock \bibinfo{title}{Specific functional connectivity patterns of middle temporal gyrus subregions in children and adults with autism spectrum disorder}.
\newblock \bibinfo{journal}{Autism Research} \bibinfo{volume}{13}, \bibinfo{pages}{410--422}.
\bibitem[{Yan et~al.(2019)Yan, Chen, Li, Castellanos, Bai, Bo, Cao, Chen, Chen, Chen et~al.}]{MDD1}
\bibinfo{author}{Yan, C.G.}, \bibinfo{author}{Chen, X.}, \bibinfo{author}{Li, L.}, \bibinfo{author}{Castellanos, F.X.}, \bibinfo{author}{Bai, T.J.}, \bibinfo{author}{Bo, Q.J.}, \bibinfo{author}{Cao, J.}, \bibinfo{author}{Chen, G.M.}, \bibinfo{author}{Chen, N.X.}, \bibinfo{author}{Chen, W.}, et~al., \bibinfo{year}{2019}.
\newblock \bibinfo{title}{Reduced default mode network functional connectivity in patients with recurrent major depressive disorder}.
\newblock \bibinfo{journal}{Proceedings of the National Academy of Sciences} \bibinfo{volume}{116}, \bibinfo{pages}{9078--9083}.
\bibitem[{Yan et~al.(2016)Yan, Wang, Zuo and Zang}]{dpabi}
\bibinfo{author}{Yan, C.G.}, \bibinfo{author}{Wang, X.D.}, \bibinfo{author}{Zuo, X.N.}, \bibinfo{author}{Zang, Y.F.}, \bibinfo{year}{2016}.
\newblock \bibinfo{title}{Dpabi: data processing \& analysis for (resting-state) brain imaging}.
\newblock \bibinfo{journal}{Neuroinformatics} \bibinfo{volume}{14}, \bibinfo{pages}{339--351}.
\bibitem[{Yeo et~al.(2011)Yeo, Krienen, Sepulcre, Sabuncu, Lashkari, Hollinshead, Roffman, Smoller, Z{\"o}llei, Polimeni et~al.}]{yeo7}
\bibinfo{author}{Yeo, B.T.}, \bibinfo{author}{Krienen, F.M.}, \bibinfo{author}{Sepulcre, J.}, \bibinfo{author}{Sabuncu, M.R.}, \bibinfo{author}{Lashkari, D.}, \bibinfo{author}{Hollinshead, M.}, \bibinfo{author}{Roffman, J.L.}, \bibinfo{author}{Smoller, J.W.}, \bibinfo{author}{Z{\"o}llei, L.}, \bibinfo{author}{Polimeni, J.R.}, et~al., \bibinfo{year}{2011}.
\newblock \bibinfo{title}{The organization of the human cerebral cortex estimated by intrinsic functional connectivity}.
\newblock \bibinfo{journal}{Journal of neurophysiology} .
\bibitem[{Yu et~al.(2024)Yu, Jin, Li, Sarwar and Xia}]{ALTER}
\bibinfo{author}{Yu, S.}, \bibinfo{author}{Jin, S.}, \bibinfo{author}{Li, M.}, \bibinfo{author}{Sarwar, T.}, \bibinfo{author}{Xia, F.}, \bibinfo{year}{2024}.
\newblock \bibinfo{title}{Long-range brain graph transformer}.
\newblock \bibinfo{journal}{Advances in Neural Information Processing Systems} \bibinfo{volume}{37}, \bibinfo{pages}{24472--24495}.
\bibitem[{Zeng et~al.(2025)Zeng, Gong, Li and Yang}]{KMGCN}
\bibinfo{author}{Zeng, X.}, \bibinfo{author}{Gong, J.}, \bibinfo{author}{Li, W.}, \bibinfo{author}{Yang, Z.}, \bibinfo{year}{2025}.
\newblock \bibinfo{title}{Knowledge-driven multi-graph convolutional network for brain network analysis and potential biomarker discovery}.
\newblock \bibinfo{journal}{Medical Image Analysis} \bibinfo{volume}{99}, \bibinfo{pages}{103368}.
\bibitem[{Zhang et~al.(2023a)Zhang, Song, Wang, Zhang, Wang, Wang and Zhang}]{9936686}
\bibinfo{author}{Zhang, H.}, \bibinfo{author}{Song, R.}, \bibinfo{author}{Wang, L.}, \bibinfo{author}{Zhang, L.}, \bibinfo{author}{Wang, D.}, \bibinfo{author}{Wang, C.}, \bibinfo{author}{Zhang, W.}, \bibinfo{year}{2023}a.
\newblock \bibinfo{title}{Classification of brain disorders in rs-fmri via local-to-global graph neural networks}.
\newblock \bibinfo{journal}{IEEE Transactions on Medical Imaging} \bibinfo{volume}{42}, \bibinfo{pages}{444--455}.
\newblock \DOIprefix\doi{10.1109/TMI.2022.3219260}.
\bibitem[{Zhang et~al.(2023b)Zhang, Wen, Cao, Yang, Zhang, Zhang, Zhu, Zaiane and Wang}]{BrianUSL}
\bibinfo{author}{Zhang, P.}, \bibinfo{author}{Wen, G.}, \bibinfo{author}{Cao, P.}, \bibinfo{author}{Yang, J.}, \bibinfo{author}{Zhang, J.}, \bibinfo{author}{Zhang, X.}, \bibinfo{author}{Zhu, X.}, \bibinfo{author}{Zaiane, O.R.}, \bibinfo{author}{Wang, F.}, \bibinfo{year}{2023}b.
\newblock \bibinfo{title}{Brainusl: Unsupervised graph structure learning for functional brain network analysis}, in: \bibinfo{booktitle}{International Conference on Medical Image Computing and Computer-Assisted Intervention}, \bibinfo{organization}{Springer}. pp. \bibinfo{pages}{205--214}.
\bibitem[{Zheng et~al.(2024)Zheng, Yu, Li, Jenssen and Chen}]{BrainIB}
\bibinfo{author}{Zheng, K.}, \bibinfo{author}{Yu, S.}, \bibinfo{author}{Li, B.}, \bibinfo{author}{Jenssen, R.}, \bibinfo{author}{Chen, B.}, \bibinfo{year}{2024}.
\newblock \bibinfo{title}{Brainib: Interpretable brain network-based psychiatric diagnosis with graph information bottleneck}.
\newblock \bibinfo{journal}{IEEE Transactions on Neural Networks and Learning Systems} .

\end{thebibliography}



\end{document}